\documentclass[10pt,twocolumn,letterpaper]{article}

\PassOptionsToPackage{table}{xcolor}
\usepackage[pagenumbers]{cvpr}
\usepackage[T1]{fontenc}
\usepackage{cuted}        
\usepackage{tcolorbox}
\tcbuselibrary{breakable,skins}
\usepackage{array}
\usepackage{tikz}
\usepackage{pgfplots}
\pgfplotsset{compat=1.18}
\usepgfplotslibrary{statistics}
\usetikzlibrary{arrows.meta, positioning, calc, shadows, fit, backgrounds}

\definecolor{cvprblue}{rgb}{0.21,0.49,0.74}
\usepackage[pagebackref,breaklinks,colorlinks,allcolors=cvprblue]{hyperref}
\hypersetup{pdftitle={How Far Can Frozen Vision-Language Models Go on Event Data? A Training-Free Study on EventBench},
            pdfauthor={Md Kamrujjaman Mobin, Md Salman Shamil, Md Adnan Arefeen}}

\newcommand{\code}[1]{\texttt{\footnotesize #1}}

\makeatletter
\renewcommand\paragraph{\@startsection{paragraph}{4}{\z@}{5pt plus 1pt minus 1pt}{-0.6em}{\normalfont\normalsize\bfseries}}
\makeatother


\definecolor{catGR}{HTML}{A9C47F}    
\definecolor{catDU}{HTML}{EDC06A}    
\definecolor{catAR}{HTML}{C3A0C8}    
\definecolor{catCR}{HTML}{93B5DE}    
\definecolor{catOCR}{HTML}{E5AE8C}   
\definecolor{inkdark}{HTML}{404040}
\definecolor{inkslate}{HTML}{4E5969}
\definecolor{pillblue}{HTML}{E4ECF9}
\definecolor{pillbeige}{HTML}{F4EBE4}
\definecolor{pillmint}{HTML}{E6F1ED}
\definecolor{evON}{RGB}{222,86,80}
\definecolor{evOFF}{RGB}{72,126,214}

\newcommand{\excol}{black}
\newcommand{\softpill}[2]{\tikz[baseline=(p.base)]\node[fill=#1,rounded corners=3.5pt,inner xsep=4pt,inner ysep=1.4pt,font=\sffamily\tiny,text=inkslate](p){#2};}
\newcommand{\excard}[7]{%
  \renewcommand{\excol}{#1}%
  \begin{tcolorbox}[enhanced, colback=#1!12, colframe=#1!12, boxrule=0pt, arc=3.5pt,
    left=4pt, right=4pt, top=2.5pt, bottom=2pt, toptitle=1.2pt, bottomtitle=1.2pt,
    colbacktitle=#1!55, coltitle=inkdark, fonttitle=\sffamily\bfseries\footnotesize,
    before skip=0pt, after skip=0pt,
    title={#2\hfill\softpill{white}{#5}\,\softpill{white}{#6 bins}\,\softpill{white}{#7}}]
  #3\par\vspace{2pt}
  {\sffamily\scriptsize\raggedright\color{inkdark}#4\par}
  \end{tcolorbox}}
\newcommand{\exq}[1]{{\bfseries Q:} #1\par\vspace{1.5pt}}
\newcommand{\exopts}[1]{#1}
\newcommand{\exopt}[2]{\hangindent=1.35em\hangafter=1 \makebox[1.35em][l]{\textcolor{inkslate}{#1.}}#2\par}
\newcommand{\exoptpred}[2]{\hangindent=1.35em\hangafter=1 \makebox[1.35em][l]{\tikz[baseline=(c.base)]\node[circle,fill=inkslate,text=white,inner sep=0pt,minimum size=1.05em,font=\sffamily\bfseries\tiny](c){#1};}{\bfseries #2}\par}
\newcommand{\exoptsgrid}[4]{{%
  \renewcommand{\exopt}[2]{\makebox[1.35em][l]{\textcolor{inkslate}{##1.}}##2}%
  \renewcommand{\exoptpred}[2]{\makebox[1.35em][l]{\tikz[baseline=(c.base)]\node[circle,fill=inkslate,text=white,inner sep=0pt,minimum size=1.05em,font=\sffamily\bfseries\tiny](c){##1};}{\bfseries ##2}}%
  \setlength{\tabcolsep}{0pt}\renewcommand{\arraystretch}{1.1}%
  \begin{tabular}{@{}p{0.5\linewidth}p{0.5\linewidth}@{}}#1&#2\\#3&#4\end{tabular}}\par}
\newcommand{\predpill}{\tikz[baseline=(p.base)]\node[fill=white,draw=inkslate!35,rounded corners=3.5pt,inner xsep=4pt,inner ysep=1.4pt,font=\sffamily\bfseries\tiny,text=inkslate](p){Our prediction};}
\newcommand{\expredtxt}[2]{\predpill~{\bfseries\color{inkslate}#2}}
\newcommand{\expredimg}[2]{\predpill\par\vspace{2pt}#2}
\newcommand{\exsep}{\par\vspace{2.5pt}}

\title{How Far Can Frozen Vision-Language Models Go on Event Data?\\A Training-Free Study on EventBench}

\author{
Md Kamrujjaman Mobin$^{1}$ \qquad Md Salman Shamil$^{2}$ \qquad Md Adnan Arefeen$^{2}$\\[4pt]
$^{1}$Shahjalal University of Science and Technology, Sylhet, Bangladesh\\
$^{2}$North South University, Dhaka, Bangladesh\\[2pt]
{\tt\small kamrujjamanmobin123@gmail.com, salman.shamil@northsouth.edu, adnan.arefeen@northsouth.edu}\\[4pt]
{\small\itshape Technical report, EBMV @ ECCV 2026 Challenge ``Event Stream Understanding with EventBench''}
}

\begin{document}
\maketitle

\begin{strip}
\centering
\vspace{-14pt}
\begin{minipage}[t]{0.655\textwidth}\vspace{0pt}
{\sffamily\bfseries\footnotesize\color{inkdark}(a) A training-free pipeline}\par\vspace{4pt}
\resizebox{\linewidth}{!}{%
\begin{tikzpicture}[
    font=\sffamily\scriptsize,
    lbl/.style={align=center, font=\sffamily\scriptsize, text=black!65},
    arr/.style={-{Stealth[length=2mm]}, line width=0.8pt, black!30},
    card/.style={draw=black!12, fill=white, rounded corners=4pt, align=center, inner sep=4pt, text=inkdark,
                 drop shadow={opacity=0.08, shadow xshift=0.5pt, shadow yshift=-0.5pt}},
    tag/.style={rounded corners=3pt, inner xsep=4pt, inner ysep=1.6pt, text=inkdark, font=\sffamily\bfseries\tiny},
    lane/.style={draw=black!14, dashed, rounded corners=6pt, inner sep=5pt}]
  \node[inner sep=0] (b2) at (0.36,0.30) {\includegraphics[width=1.9cm]{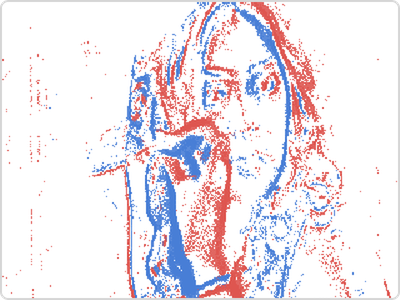}};
  \node[inner sep=0] (b1) at (0.18,0.15) {\includegraphics[width=1.9cm]{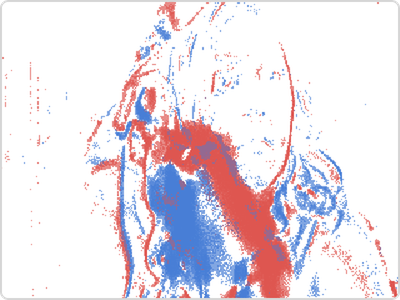}};
  \node[inner sep=0] (b0) at (0,0) {\includegraphics[width=1.9cm]{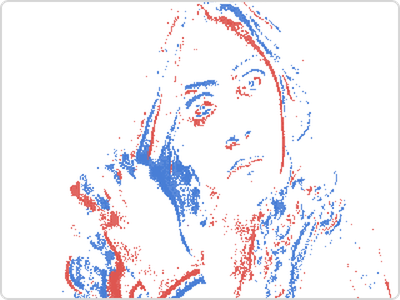}};
  \node[lbl, below=2pt of b0] (bl) {raw event bins\\(8--32 per clip)};
  \node[inner sep=0] (grid) at (4.55,0.10) {\includegraphics[width=3.7cm]{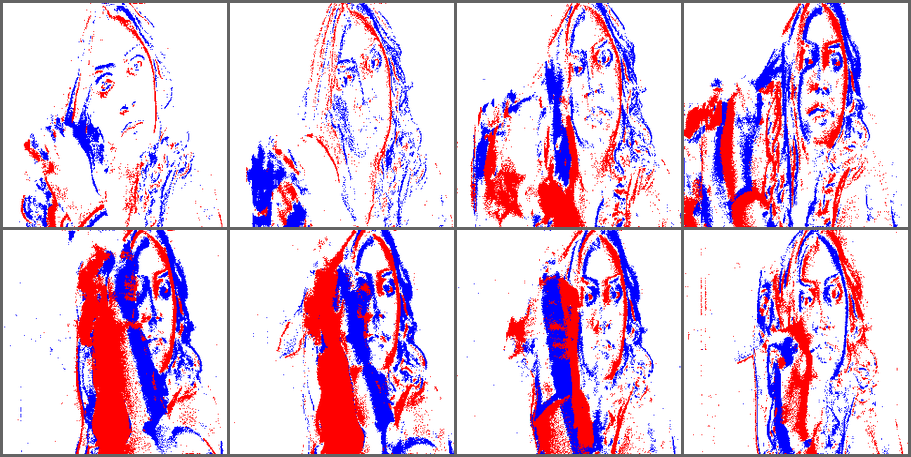}};
  \node[lbl, below=2pt of grid] (gl) {one grid image, $k$ cells in time order\\(the exact model input)};
  \draw[arr] ($(b2.east)+(0.10,-0.22)$) -- node[lbl, above] {select} node[lbl, below] {crop} ($(grid.west)+(-0.10,0)$);
  \node[card, minimum width=2.45cm, minimum height=1.45cm] (vlm) at (8.45,0.10)
    {{\bfseries\small Frozen VLM}\\[1pt] Qwen3.5 9B\\[1pt]
     \textcolor{black!50}{+ category prompt}};
  \draw[arr] ($(grid.east)+(0.10,0)$) -- (vlm.west);
  \node[tag, fill=white, draw=inkslate!40, text=inkslate, font=\sffamily\bfseries\scriptsize] (ans) at (10.75,0.10) {Answer};
  \draw[arr] (vlm.east) -- (ans.west);
  \node[tag, fill=catGR!55, anchor=south west] (t1) at ($(b0.north west)+(-0.05,0.42)$) {Gesture};
  \node[tag, fill=catDU!55, anchor=west] (t2) at ($(t1.east)+(0.07,0)$) {Detailed};
  \node[tag, fill=catCR!55, anchor=west] (t3) at ($(t2.east)+(0.07,0)$) {Causal};
  \node[tag, fill=catOCR!55, anchor=west] (t4) at ($(t3.east)+(0.07,0)$) {Event OCR};
  \begin{scope}[on background layer]
    \node[lane, fit=(t1)(b2)(bl)(gl)(vlm)(ans)] (L1) {};
  \end{scope}
  \node[tag, fill=catAR!55, anchor=west] (artag) at ($(L1.south west)+(0.15,-0.62)$) {Action};
  \node[card, anchor=west] (all) at ($(artag.east)+(0.45,0)$) {all bins, no selection or grid};
  \node[card, anchor=west] (egpt) at ($(all.east)+(0.55,0)$) {{\bfseries EventGPT-Plus-2B}, native event encoder};
  \node[tag, fill=white, draw=inkslate!40, text=inkslate, font=\sffamily\bfseries\scriptsize] (ans2) at (10.75,0|-artag) {Answer};
  \draw[arr] (artag.east) -- (all.west);
  \draw[arr] (all.east) -- (egpt.west);
  \draw[arr] (egpt.east) -- (ans2.west);
  \begin{scope}[on background layer]
    \node[lane, fit=(artag)(all)(egpt)(ans2)] {};
  \end{scope}
\end{tikzpicture}}
\end{minipage}\hfill
\begin{minipage}[t]{0.325\textwidth}\vspace{0pt}
{\sffamily\bfseries\footnotesize\color{inkdark}(b) Test accuracy, no training data}\par\vspace{2pt}
\begin{tikzpicture}
\begin{axis}[
    xbar, bar width=7pt, scale only axis, width=0.6\linewidth, height=3.75cm,
    xmin=0, xmax=100, enlarge y limits=0.13,
    symbolic y coords={OCR,Gesture,Detailed,Causal,Action},
    ytick={OCR,Gesture,Detailed,Causal,Action}, yticklabels={Event OCR,Gesture,Detailed,Causal,Action},
    y tick label style={font=\sffamily\scriptsize}, point meta=x,
    xtick={0,25,50,75,100}, x tick label style={font=\sffamily\tiny},
    xlabel={accuracy (\%)}, xlabel style={font=\sffamily\tiny, yshift=3pt},
    axis lines*=left, xmajorgrids, major grid style={black!8},
    nodes near coords={\pgfmathprintnumber[fixed,precision=2,fixed zerofill]{\pgfplotspointmeta}},
    nodes near coords style={font=\sffamily\tiny\bfseries, text=inkdark, anchor=west},
    bar shift=0pt,
]
\addplot[fill=catAR!80, draw=none] coordinates {(80.27,Action)};
\addplot[fill=catCR!80, draw=none] coordinates {(77.22,Causal)};
\addplot[fill=catDU!80, draw=none] coordinates {(66.46,Detailed)};
\addplot[fill=catGR!80, draw=none] coordinates {(54.70,Gesture)};
\addplot[fill=catOCR!80, draw=none] coordinates {(34.57,OCR)};
\draw[dashed, black!40] ([yshift=-8pt]axis cs:25,OCR) -- ([yshift=8pt]axis cs:25,Action)
   node[above, font=\sffamily\tiny, text=black!55] {chance};
\draw[densely dotted, evON!80, line width=0.7pt] ([yshift=-8pt]axis cs:62.28,OCR) -- ([yshift=8pt]axis cs:62.28,Action)
   node[above, font=\sffamily\tiny\bfseries, text=evON!85!black] {overall 62.28};
\end{axis}
\end{tikzpicture}
\end{minipage}
\captionsetup{hypcap=false}

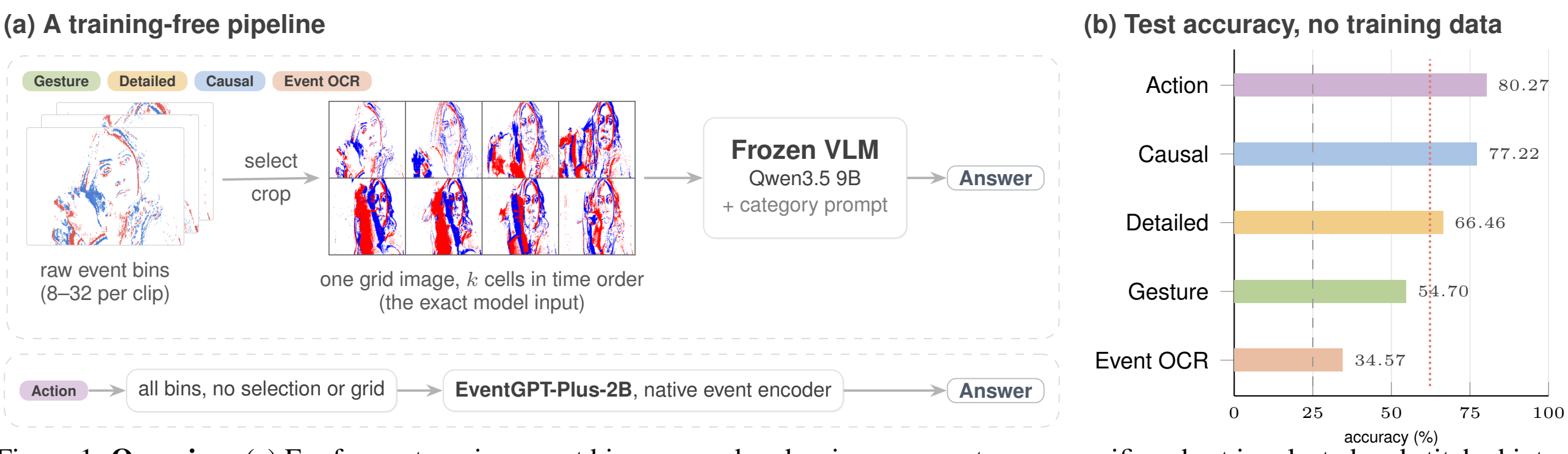
\captionof{figure}{\textbf{Overview.} (a) For four categories, event bins are rendered as images, a
category-specific subset is selected and stitched into one grid (shown: the exact model input for a
detailed-understanding clip of a woman adjusting her hair, used only in this figure), and a frozen vision-language model answers from it;
action recognition passes every bin to the organizers' EventGPT-Plus-2B. (b) Accuracy on the
held-out test set; chance is 25\% for the four multiple-choice tasks. The overall
accuracy pools correct answers across all 2{,}084 questions. \textbf{Without any training data,
the system reaches 62.28\% overall accuracy, rank 9 of 27.}}
\captionsetup{hypcap=true}
\label{fig:teaser}
\vspace{2pt}
\end{strip}

\begin{abstract}
We present a training-free system for the EBMV @ ECCV 2026 challenge ``Event Stream
Understanding with EventBench'' (General Understanding track),
which evaluates vision-language models on five event-camera understanding tasks. We
use none of the challenge's training data: our goal is to measure how far
off-the-shelf, RGB-pretrained models can go on event data through representation and
prompting alone. For four of the five categories (gesture recognition, detailed
understanding, causal reasoning, and event OCR) we render event bins as red/ON,
blue/OFF frames, select a category-specific subset of bins, stitch them into a single
grid image, and query the same frozen Qwen3.5 9B vision-language model with a
category-specific prompt for all four tasks. The fifth
category, action recognition, feeds every bin directly into the organizers'
EventGPT-Plus-2B. The system reaches 62.28\% overall accuracy across 2{,}084
test questions (rank 9/27), led by
action recognition and causal reasoning (80.27\%, 77.22\%) and limited most by gesture
recognition and event OCR (54.70\%, 34.57\%). We visualize test samples from every
category together with the exact images the model receives, score six alternative
representations against the evaluation server, and analyze the full test set
post hoc. That analysis shows that our bounding-box crop almost never tightened around
content, so event OCR text reached the model at roughly one fifth of its horizontal
resolution, and that detailed-understanding predictions are skewed toward the first
two answer options.
\end{abstract}

\section{Introduction}
\label{sec:intro}
Event cameras are neuromorphic sensors that asynchronously report per-pixel brightness
changes rather than capturing synchronous frames, giving them microsecond-scale temporal
resolution, very high dynamic range ($>$120\,dB), low power draw, and robustness to motion
blur that conventional frame-based cameras lack~\cite{gallego}. These properties make them
attractive for robotics, autonomous driving, AR/VR, and other latency- or
power-constrained settings. However, the sensor's native output is a sparse, asynchronous
stream of $(x, y, t, p)$ events rather than a dense image, and it is not directly consumable by
the frame-based vision models, including modern multimodal large language models (MLLMs),
that dominate the field. How to expose event data to a model pretrained on conventional
images is itself an open representational question, and is the problem we address here.

The EventBench benchmark~\cite{eventbench}\footnote{Project page:
\url{https://xdusyl.github.io/eventbench.github.io/}; data:
\url{https://huggingface.co/datasets/XduSyL/EventBench}.} evaluates MLLM understanding of event
streams. Its challenge at the EBMV workshop at ECCV 2026~\cite{ebmv2026}, ``Event Stream
Understanding with EventBench''~\cite{eventbenchchallenge} (the General Understanding track),
covers five tasks: gesture recognition, detailed
understanding, action recognition, causal reasoning (each 4-way multiple choice), and event
OCR (free text), with 2{,}084 test questions in total (Table~\ref{tab:data},
Figure~\ref{fig:examples}).

The organizers' own reference baseline, EventGPT~\cite{eventgpt}, is a trained model, as are its recent successors~\cite{eventvl,letus,eventflash}.
Rather than train a model ourselves, we ask a narrower question:
given a fixed, general-purpose, off-the-shelf vision-language model with a
single-image-per-call interface, how much of the task can be solved purely by choosing what
to show it and how to ask, without touching the challenge's training data at all
(Figure~\ref{fig:teaser})? Our contributions are:
\begin{itemize}[leftmargin=*, itemsep=1pt, topsep=2pt]
\item a training-free grid-of-bins representation with task-specific bin selection and
category-specific prompts, given verbatim (Section~\ref{sec:method},
Appendix~\ref{app:prompts});
\item a visual walk-through of test samples from every category, paired with the exact image
the model receives (Figures~\ref{fig:examples} and~\ref{fig:inputs},
Appendix~\ref{app:examples});
\item scored comparisons of six alternative representations and bin-selection strategies
against the official evaluation server (Section~\ref{sec:additional-experiments});
\item a post-hoc analysis over the full test set showing that our crop almost never
tightened around content, which leaves event OCR text heavily downsampled
(Section~\ref{sec:analysis}), and that detailed-understanding answers are skewed toward the first
options (Section~\ref{sec:analysis}).
\end{itemize}

\section{Related Work}
\label{sec:related}
\paragraph{Event-based vision.} Classical event-vision research, surveyed comprehensively by
Gallego et al.~\cite{gallego}, spans low-level tasks (optical flow, depth, tracking) built
on hand-designed event representations (time surfaces, voxel grids, event frames) and
specialized asynchronous algorithms, predating the current wave of MLLM-based approaches.
Our per-bin rendering (Section~\ref{sec:render}) is one such simple, classical
representation (an event frame), chosen deliberately for compatibility with an
off-the-shelf image-only model rather than for novelty. Recent work also pushes zero-shot
generalization outside the MLLM paradigm: Match-Any-Events~\cite{matchanyevents} learns a
motion-robust event feature matcher that generalizes across datasets \emph{without any
target-domain fine-tuning}, the same no-adaptation constraint we operate under, applied
to wide-baseline correspondence rather than question answering.

\paragraph{Event-based gesture and motion.} Gesture and action recognition interpret human
motion from events, a problem with substantial dedicated literature that we bypass by
using a single generic grid representation. EgoEvGesture~\cite{egoevgesture} designs a
gesture-specific architecture around a parameter-free \emph{Bins-Temporal Shift Module}
that fuses information across event bins inside the network, in contrast to our approach
of fusing bins visually (as separate grid cells) and leaving the fusion to the frozen
VLM's attention. Wang et al.~\cite{motionfield} instead predict a \emph{continuous-time}
human motion field directly from raw events, avoiding discrete-time frame sampling
altogether; we return to this contrast in Section~\ref{sec:mainresults} when discussing why
our 8-bin sampling likely under-resolves fine-grained gesture motion.

\paragraph{Event-based MLLMs and benchmarks.} EventGPT~\cite{eventgpt} established the recipe
most event MLLMs follow: an event encoder, a spatio-temporal aggregator, and an event-language
adapter in front of an LLM, trained in three stages on synthetic and real event-text data.
Later models scale this recipe. EventVL~\cite{eventvl} annotates about 1.4 million
event-image/video-text pairs, LET-US~\cite{letus} compresses long event streams for
event-text understanding, and EventFlash~\cite{eventflash} sparsifies spatio-temporal tokens
and builds the 500k-instruction EventMind dataset. Other models fuse events with RGB frames:
for reasoning under extreme illumination~\cite{eventmllm}, for general scene
understanding~\cite{revlm}, and for driving~\cite{eventdrive}. For evaluation,
EvQA~\cite{evqa} provides 1,000 real-world event question-answer pairs and studies frame
reconstruction as a bridge to frame-based MLLMs, and EventBench~\cite{eventbench} covers
eight tasks, five of which form this challenge track. Every model above adapts parameters to
event data (Table~\ref{tab:related}). We instead keep a general-purpose VLM fixed for four
of the five categories, and use EventGPT's released checkpoint as-is for the fifth
(Section~\ref{sec:action}).

\paragraph{Aligning events with pretrained models.} A lighter line of work reuses image-text
models. EventCLIP~\cite{eventclip} converts events into 2D grid representations for CLIP and
adds a feature adapter for few-shot recognition, EventBind~\cite{eventbind} generalizes CLIP's
encoders to events for open-world recognition, and GEP~\cite{gep} aligns an event encoder to a
frozen vision foundation model before generative pretraining. Closest to us,
LLM-EvRep~\cite{llmevrep} trains a self-supervised generator of LLM-compatible event images
and reports gains over E2VID reconstructions~\cite{e2vid} when GPT-4o reads them. Our grid
rendering is the untrained end of this spectrum: a fixed, hand-designed image. In the other
direction, EventSTU~\cite{eventstu} borrows the change-triggered principle of event cameras
to make video LLMs cheaper without training, and EP-VLM~\cite{epvlm} uses event-derived
motion priors to sparsify the RGB input of a VLM.

\begin{table}[t]
\centering
\footnotesize
\setlength{\tabcolsep}{3pt}
\begin{tabular}{@{}l >{\raggedright\arraybackslash}p{3.45cm} l@{}}
\toprule
\textbf{Method} & \textbf{How events enter the model} & \textbf{Event training} \\
\midrule
EventGPT~\cite{eventgpt} & event encoder, aggregator, adapter & \checkmark{} three stages \\
EventVL~\cite{eventvl} & event spatio-temporal representation & \checkmark{} 1.4M pairs \\
LET-US~\cite{letus} & compressed long-stream features & \checkmark{} two stages \\
EventFlash~\cite{eventflash} & sparsified spatio-temporal tokens & \checkmark{} curriculum \\
Event-MLLM~\cite{eventmllm} & event features fused with RGB & \checkmark \\
RE-VLM~\cite{revlm} & parallel RGB and event encoders & \checkmark{} progressive \\
EventCLIP~\cite{eventclip} & 2D event grids into CLIP & adapter (few-shot) \\
LLM-EvRep~\cite{llmevrep} & learned event images, frozen GPT-4o & generator only \\
\midrule
\rowcolor{catCR!22} \textbf{Ours} & rendered bins stitched into one grid & \textbf{none} (frozen VLM) \\
\bottomrule
\end{tabular}
\caption{\textbf{Positioning against recent event-language models.} Descriptions follow each
paper's abstract. All prior models adapt parameters to event data; for four of five categories
our system adapts nothing and changes only the rendered input and the prompt.}
\label{tab:related}
\end{table}

\paragraph{Prompting non-natural-image modalities.} A separate line of work asks how far a
frozen vision model can go on data it was never trained on, purely by choosing how that
data is rendered as an image. For example, VisionTS~\cite{visionts} reformats 1-D time series
as 2-D image patches for a masked-autoencoder vision model to forecast zero-shot. Our work
applies the same philosophy to event streams: no modality-specific encoder, only a
rendering choice.

\paragraph{Event-based OCR.} Our event OCR representation is informed by Brander et
al.~\cite{foveated}, who propose an eye-gaze-foveated, deep-binary-reconstructed event OCR
pipeline for smart glasses, feeding the reconstructed frame to an MLLM for reading. We
attempted two of their three stages without any training budget: \emph{foveation}, via
per-bin cropping (Section~\ref{sec:crop}) instead of gaze tracking, which as we show
(Section~\ref{sec:analysis}) was
largely ineffective as implemented; and the final \emph{VLM-OCR} stage, via an
off-the-shelf vision-language model. We do not attempt their trained binary-reconstruction
network; our grid-of-bins representation is a training-free substitute for that stage,
asking the model to reconstruct characters by cross-referencing multiple raw cells instead
of reading one clean reconstructed frame. The tradeoff versus all of these directions is
direct: no training cost or labeled-data requirement, at the cost of being bounded by
what a general-purpose model can already recognize in a synthetic red/blue rendering it
was never trained on.

\section{Benchmark and Data}
\label{sec:data}
\begin{figure}[t]
\centering
\input{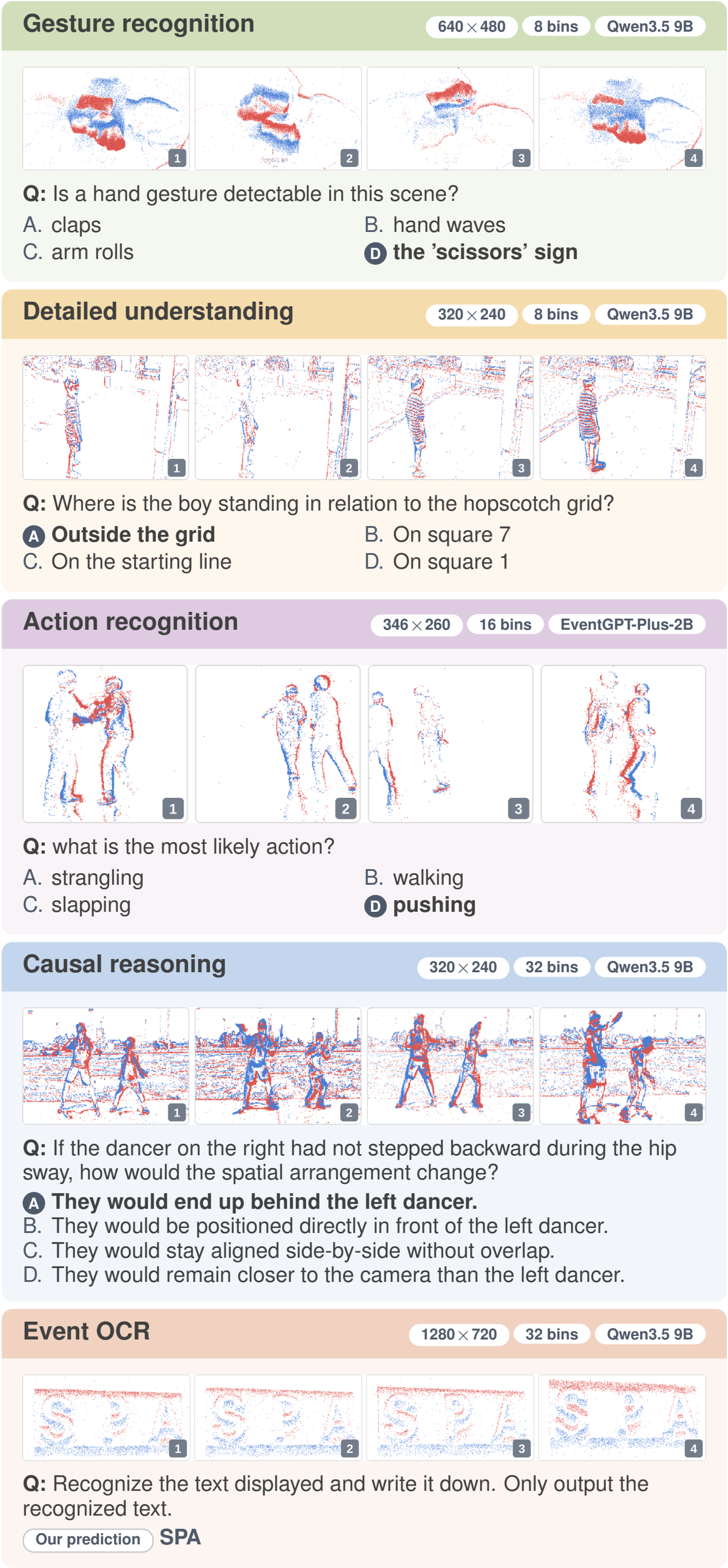}
\caption{\textbf{One test sample per category}, four bins in time order (display rendering,
\textcolor{evON}{red = ON}, \textcolor{evOFF}{blue = OFF}). Labels are withheld, so the filled
circle marks \emph{our prediction}, not the answer. Random samples: Appendix~\ref{app:examples};
exact model inputs: Figure~\ref{fig:inputs}.}
\label{fig:examples}
\end{figure}

Each test question references one NPZ file holding 8 to 32 event bins, each a list of
pixel coordinates and polarities, with no per-event timestamp beyond the bin order. Table~\ref{tab:data} summarizes the test set, computed by us
from the released files, and Figure~\ref{fig:examples} shows one sample per category.
Ground truth is withheld; the challenge's evaluation server~\cite{eventbenchchallenge} scores
exact match per category and ranks
teams by accuracy pooled over all questions.

The five categories differ far more than their shared question format suggests. Detailed
understanding and causal reasoning are drawn mainly from RGB video datasets converted to
events with the V2E simulator~\cite{v2e,eventbench}, producing dense bins (a median of 29k and 40k
events per bin) in which the whole scene is visible. Gesture, action, and OCR clips come
mainly from real event cameras and are much sparser (a median of 3.5k events per bin for
action recognition), so only moving edges appear. Resolution also varies within a
category: 400 of the 894 gesture clips are recorded at a lower resolution than the
remaining 494, and the event OCR recordings are $1280{\times}720$, twelve times the pixel
count of a $320{\times}240$ clip. The OCR source, EventSTR, contains both Chinese and English
text~\cite{eventstr}, and 124 of our 162 OCR outputs contain Chinese characters, so most
targets are evidently Chinese (Appendix~\ref{app:ocrout}).

\begin{table*}[t]
\centering
\footnotesize
\setlength{\tabcolsep}{4.5pt}
\begin{tabular}{@{}l r l p{6.1cm} l c r@{}}
\toprule
\textbf{Category} & \textbf{Qs} & \textbf{Answer} & \textbf{Source datasets (per EventBench~\cite{eventbench})} & \textbf{Resolution} & \textbf{Bins/clip} & \textbf{Events/bin} \\
\midrule
{\color{catGR}\rule{5pt}{5pt}}\ Gesture recognition & 894 & 4-way MC & EB-HandGesture~\cite{ebhandgesture}, EHWGesture~\cite{ehwgesture} & $640{\times}480$ / lower & 8 & 10.2k \\
{\color{catDU}\rule{5pt}{5pt}}\ Detailed understanding & 483 & 4-way MC & Kinetics~\cite{kinetics}, SportsSloMo~\cite{sportsslomo}, MotionBench~\cite{motionbench}, PLM-Data~\cite{perceptionlm}, PE-Data~\cite{perceptionencoder}, DSEC~\cite{dsec}, N-ImageNet~\cite{nimagenet} & $320{\times}240$ & 16 & 29.1k \\
{\color{catAR}\rule{5pt}{5pt}}\ Action recognition & 365 & 4-way MC & DailyDVS-200~\cite{dailydvs}, Bullying10K~\cite{bullying10k}, UCF-101~\cite{ucf101}, HARDVS~\cite{hardvs} & $320{\times}240$, $346{\times}260$ & 16 & 3.5k \\
{\color{catCR}\rule{5pt}{5pt}}\ Causal reasoning\textsuperscript{a} & 180 & 4-way MC & ActivityNet~\cite{activitynet}, Charades~\cite{charades}, MotionSight~\cite{motionsight}, WebVid-10M~\cite{webvid} & $320{\times}240$ & 32 & 39.6k \\
{\color{catOCR}\rule{5pt}{5pt}}\ Event OCR & 162 & free text & EventSTR~\cite{eventstr} & $1280{\times}720$ & 8 & 15.5k\textsuperscript{b} \\
\bottomrule
\end{tabular}
\caption{\textbf{Test set of the challenge}~\cite{eventbenchchallenge}. Source datasets are those EventBench
lists for each task; all other columns are measured by us on the released test files (bins
per clip and events per bin are medians; every category ranges from 8 to 32 bins per clip).
\textsuperscript{a}The official identifier is spelled \code{casual\_reasoning}; its 180
questions share 104 clips. \textsuperscript{b}15.6\% of event OCR bins are empty.}
\label{tab:data}
\end{table*}

\section{Method}
\label{sec:method}
\paragraph{Overview.} Figure~\ref{fig:teaser}(a) summarizes the system. For four categories, each clip's bins are rendered as images (Section~\ref{sec:render}), a category-specific subset is selected (Section~\ref{sec:binsel}) and stitched into one grid (Section~\ref{sec:crop}), and a frozen vision-language model answers from the grid and a short prompt (Section~\ref{sec:inference}). Action recognition instead feeds all bins to EventGPT-Plus-2B (Section~\ref{sec:action}).

\subsection{Event-to-Image Rendering}
\label{sec:render}
Each bin is rendered independently against a white background: OFF events ($p{=}0$,
brightness decrease) are colored blue, ON events ($p{=}1$) red, and unfired pixels remain
white. Formally, for a bin $b$ containing events $\{(x_i, y_i, p_i)\}$ and an all-white
frame $F \in \{0,\dots,255\}^{H\times W\times 3}$, each event sets a single pixel:
\begin{equation}
F[\hat{y}_i, \hat{x}_i] =
\begin{cases}
(0, 0, 255) & p_i = 0 \ \text{(OFF)} \\
(255, 0, 0) & p_i = 1 \ \text{(ON)}
\end{cases}
\end{equation}
where $\hat{x}_i = \mathrm{clip}(x_i, 0, W{-}1)$ and $\hat{y}_i = \mathrm{clip}(y_i, 0,
H{-}1)$. Clipping is required because raw event coordinates occasionally exceed the
declared sensor bounds by a few pixels in the provided data. The canvas size $H{\times}W$ is
fixed per category (Table~\ref{tab:config}).

\subsection{Bin Selection}
\label{sec:binsel}
A raw file typically contains more bins than are useful for a single stitched image. We
implement three selection strategies. \textbf{Temporal uniform} selection divides the sequence
into $k$ equal segments and keeps the busiest bin per segment; it is used for gesture
recognition, detailed understanding, and causal reasoning. \textbf{Event count} selection keeps
the top-$k$ bins by raw event count, in temporal order; it is used for event OCR, since
the busiest bins correspond to the fastest relative motion between camera and text and
therefore the densest strokes. A coverage/polarity-balance quality score, tried first for
OCR, favored motion coverage over stroke density and performed worse.
\textbf{All above threshold} selection keeps every bin above a data-driven floor (10th percentile
of per-bin event counts, sampled from $\sim$30 files) and lets the grid grow to fit; we
evaluated it for detailed understanding (Section~\ref{sec:additional-experiments}) but it
did not outperform temporal uniform selection.

\subsection{Cropping and Grid Stitching}
\label{sec:crop}
\begin{figure*}[t]
\centering
\newcommand{\inpanel}[4]{%
  \begin{minipage}[t]{0.49\textwidth}\centering
  {\sffamily\scriptsize\color{inkdark}\textcolor{#1}{\rule{6pt}{6pt}}\ \textbf{#2}}\\[3pt]
  #3
  \end{minipage}}
\inpanel{catGR}{(a) Gesture recognition: $k{=}8$, shared crop}{\includegraphics[width=0.92\linewidth]{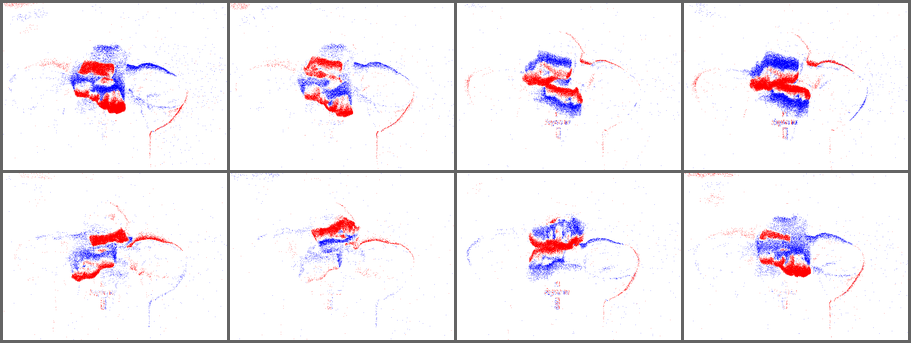}}{0.92}\hfill
\inpanel{catDU}{(b) Detailed understanding: $k{=}8$, per-bin crop}{\includegraphics[width=0.92\linewidth]{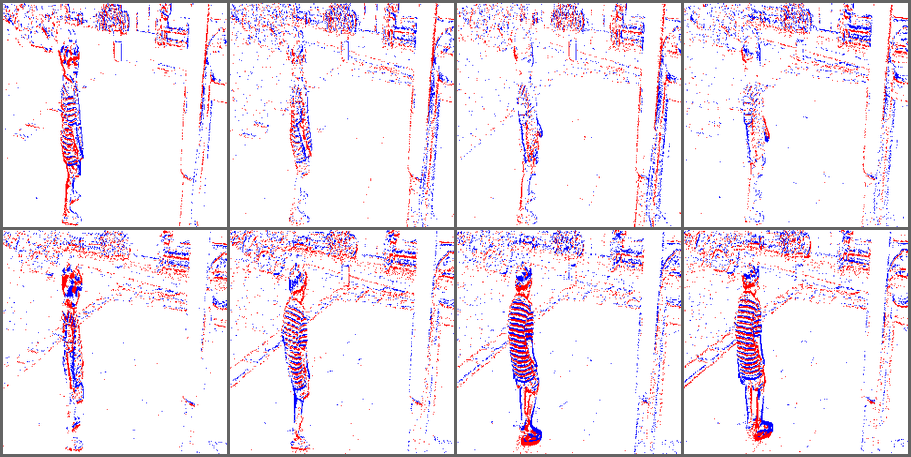}}{0.92}\\[8pt]
\inpanel{catCR}{(c) Causal reasoning: $k{=}6$, shared crop}{\includegraphics[width=0.72\linewidth]{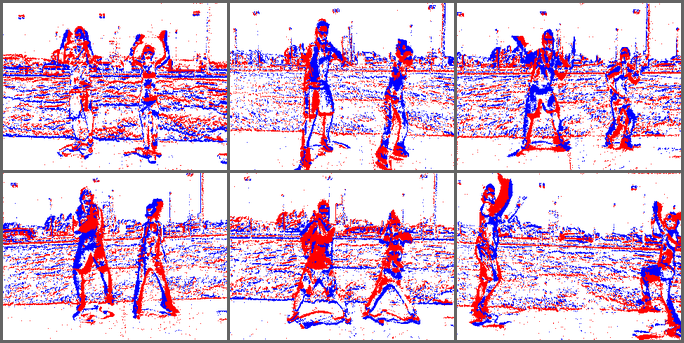}}{0.72}\hfill
\inpanel{catOCR}{(d) Event OCR: top-8 bins by event count, per-bin crop}{\includegraphics[width=0.92\linewidth]{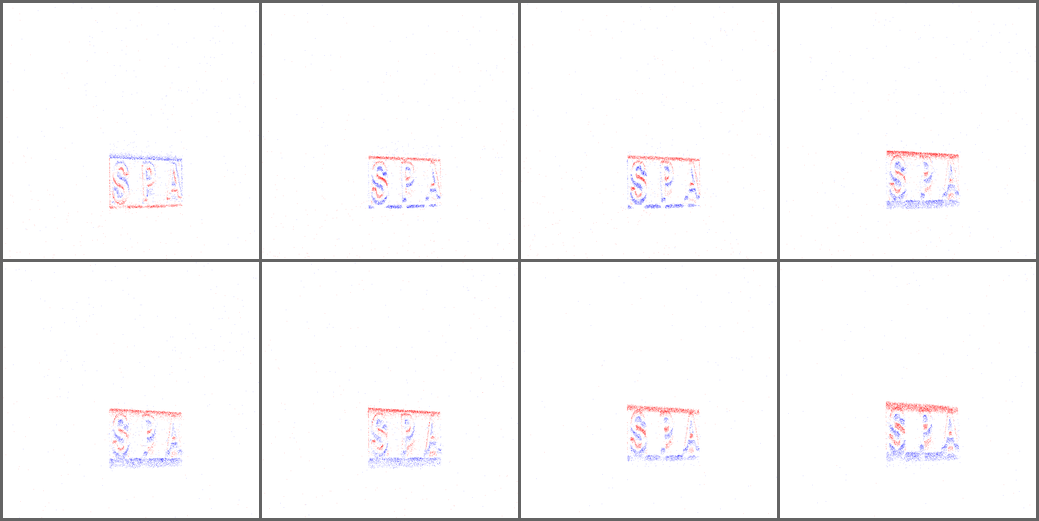}}{0.92}
\caption{\textbf{What the model actually sees.} The exact images our graded pipeline sent to the model for the four grid-category
samples of Figure~\ref{fig:examples}, regenerated from the test files with the submission's
selection, crop, and resizing settings. Note how little the crops remove: the gesture clip is
cropped only to the hand-and-arm region already filling the frame, and the event OCR text
(``SPA'') occupies a small part of each $1280{\times}720$ bin squeezed into a square cell.}
\label{fig:inputs}
\end{figure*}

The vision-language model is queried with exactly one image per call, so selected bins are
stitched into a single grid. Cells are placed left-to-right, top-to-bottom in ascending
temporal order, separated by thin gray lines (Figure~\ref{fig:inputs}). We implemented two
crop modes, both based on the min/max extent of event coordinates plus fixed padding:
\textbf{shared}, one box covering all events in the file, applied to every cell (8\%
padding, 64\,px minimum), which keeps scale consistent across cells; and
\textbf{per-bin}, an independent box per cell (12\% padding for event OCR), resized into
a square cell. Table~\ref{tab:config} lists the per-category configuration, and Section~\ref{sec:analysis} measures how much these crops removed in practice.

\begin{table}[t]
\centering
\footnotesize
\setlength{\tabcolsep}{2.2pt}
\begin{tabular}{@{}llcccl@{}}
\toprule
\textbf{Category} & \textbf{Model} & \textbf{Canvas} & \textbf{k} & \textbf{Crop} & \textbf{Select.} \\
\midrule
\rowcolor{catGR!22} Gesture & Qwen3.5 9B & $640{\times}480$ & 8 & shared & uniform \\
\rowcolor{catDU!22} Detailed & Qwen3.5 9B & $320{\times}240$ & 8 & per-bin & uniform \\
\rowcolor{catAR!22} Action & EventGPT-Plus-2B & $346{\times}260$ & all & none & none \\
\rowcolor{catCR!22} Causal & Qwen3.5 9B & $320{\times}240$ & 6 & shared & uniform \\
\rowcolor{catOCR!22} Event OCR & Qwen3.5 9B & $1280{\times}720$ & 8 & per-bin & count \\
\bottomrule
\end{tabular}
\caption{\textbf{Per-category configuration} of the graded submission (uniform = temporal
uniform selection, count = event count selection). Action recognition passes
every bin to EventGPT-Plus-2B's native encoder rather than using the grid pipeline
(Section~\ref{sec:action}). The four grid-based categories use the same frozen 9B model.}
\label{tab:config}
\end{table}

\subsection{Action Recognition: A Different Pipeline}
\label{sec:action}
Action recognition's graded predictions come from the organizers' own EventGPT-Plus-2B
reference model~\cite{eventgpt}, run locally as released, without fine-tuning. Every bin in the file is rendered with the same red/blue/white encoding as
Section~\ref{sec:render}, and the frames are passed as a stack directly into
EventGPT-Plus-2B's native multi-frame event encoder: there is no bin selection, cropping,
or grid stitching. The model's default conversation template is used with the literal
question text; no custom system prompt is applied. This pipeline predates the grid
approach and was never replaced for this category; at 80.27\% (Table~\ref{tab:results})
it is our highest-scoring category.

\subsection{Inference and Prompting}
\label{sec:inference}
The other four categories are queried through an OpenAI-compatible chat-completions
endpoint serving the open-weight Qwen3.5 9B model~\cite{qwen35}, with no access to weights or logits beyond the
returned text. The same model serves gesture recognition, detailed understanding,
causal reasoning, and event OCR; only the inputs and category prompts differ.
No model is fine-tuned or otherwise updated. Requests use
greedy decoding (temperature 0) with a 2{,}048-token output limit; thinking
mode must also be disabled, since this variant otherwise spends its
full token budget on internal reasoning and returns an empty response. System prompts are
short and category-specific, describing the color encoding and temporal ordering of the
grid; Appendix~\ref{app:prompts} gives them verbatim. Multiple-choice prompts require the
response to end with \code{Answer: X}, which is extracted by a three-stage
regular-expression parser (exact match, final-line letter, parenthesized letter).

\paragraph{Model selection and parameter-cap compliance.} \label{sec:compliance}
The challenge limits evaluated models to at most 10B parameters. All graded predictions
were obtained with models below that threshold: Qwen3.5 9B for the four grid-based
categories, including event OCR, and EventGPT-Plus-2B for action recognition.
Thus, the reported configurations satisfy the model-size constraint. Using the same
9B backbone for the four grid-based categories also avoids a model-size difference
between our adopted OCR configuration and the OCR alternatives evaluated below.

\section{Experiments}
\label{sec:results}

\subsection{Setup and Implementation Details}
\label{sec:impl}
Scores are exact-match accuracy from the official evaluation server (Section~\ref{sec:data}).

\paragraph{Compute and data handling.} All runs used Kaggle sessions with a 12-hour limit, so the larger
categories (gesture recognition: 894 samples, detailed understanding: 483) were split into
2--3 non-overlapping parts run as separate sessions and merged afterward. For the grid
pipeline, rendering, cropping, and stitching are CPU-bound NumPy/Pillow operations and
inference runs on the endpoint provider's infrastructure; EventGPT-Plus-2B for action
recognition ran locally on a single NVIDIA Tesla P100 GPU (16\,GB). Predictions are written
to disk after every sample so that an interrupted session can resume without recomputing
completed samples. Because long runs left a small number of samples empty after their
retry budget was exhausted (Section~\ref{sec:limits}), we merged repeated runs of the same
category with a non-empty-answer-wins rule, verified after every merge against the official
per-category sample counts so that no sample was dropped or duplicated.

\subsection{Main Results}
\label{sec:mainresults}
Table~\ref{tab:results} reports our final leaderboard scores. The overall 62.28\%
is pooled exact-match accuracy across the 2{,}084 questions, rather than the
unweighted mean of the five category accuracies. For reference, random
guessing on the four-way multiple-choice categories would score about 25\%. Gesture
recognition and event OCR were the weakest categories despite receiving the most
representation-engineering iteration. For gesture recognition (54.70\%), we hypothesize
that the main limit is temporal: gestures are distinguished by fine-grained hand
trajectory, which an 8-frame sample likely under-resolves compared with action
recognition's larger-scale body motion (compare the gesture and action rows of
Figure~\ref{fig:examples}). This is the information loss that dedicated event-based motion
architectures, such as continuous-time motion fields~\cite{motionfield} and gesture-specific
bin-fusion networks~\cite{egoevgesture}, are built to avoid. For event OCR (34.57\%),
Section~\ref{sec:analysis} identifies a concrete, spatial cause: the text reached the model
heavily downsampled. Neither explanation was confirmed with a controlled ablation before
the test server closed. The gap to the top leaderboard entry (90.69\% average) is largest
in these two categories, suggesting the largest remaining gains lie in gesture- and
OCR-specific representation choices rather than in the shared pipeline.

\begin{table}[t]
\centering
\small
\setlength{\tabcolsep}{4pt}
\begin{tabular}{@{}llr@{}}
\toprule
\textbf{Category} & \textbf{Model} & \textbf{Score} \\
\midrule
\rowcolor{catGR!22} Gesture recognition & Qwen3.5 9B & 54.70\% \\
\rowcolor{catDU!22} Detailed understanding & Qwen3.5 9B & 66.46\% \\
\rowcolor{catAR!22} Action recognition & EventGPT-Plus-2B & 80.27\% \\
\rowcolor{catCR!22} Causal reasoning & Qwen3.5 9B & 77.22\% \\
\rowcolor{catOCR!22} Event OCR & Qwen3.5 9B & 34.57\% \\
\midrule
\multicolumn{2}{@{}l}{\textbf{Overall (rank 9/27)}} & \textbf{62.28\%} \\
\bottomrule
\end{tabular}
\caption{\textbf{Final per-category accuracy} (exact match) on the held-out test set.
The overall score pools predictions across all 2{,}084 questions, not categories.
All models satisfy the challenge's 10B parameter cap (Section~\ref{sec:compliance}).}
\label{tab:results}
\end{table}

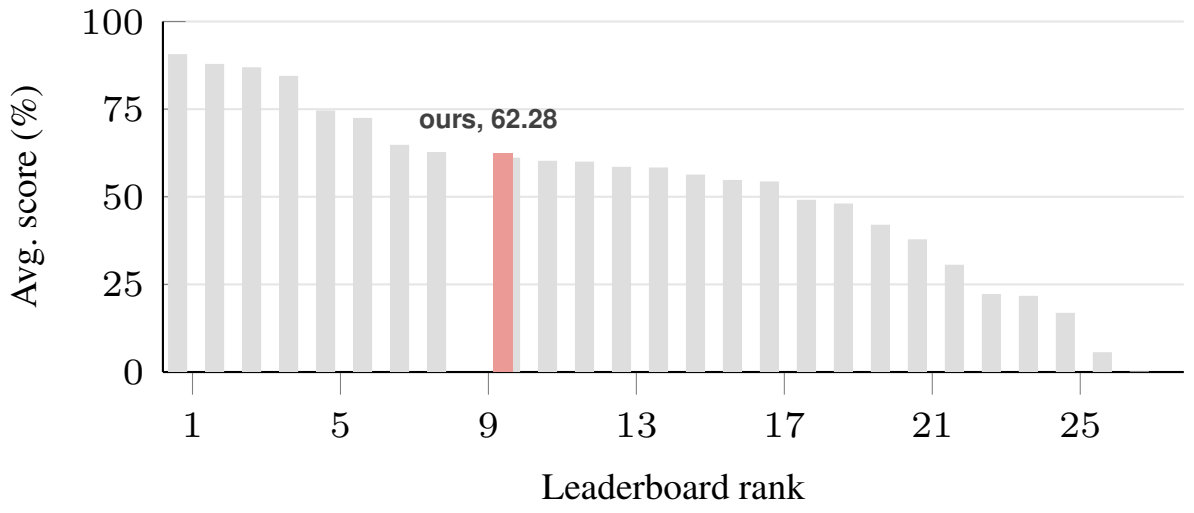
\begin{figure}[t]
\centering
\begin{tikzpicture}
\begin{axis}[
    width=\linewidth, height=3.9cm,
    ybar, bar width=3.6pt,
    ymin=0, ymax=100, xmin=0.2, xmax=27.8,
    ylabel={Avg.\ score (\%)}, ylabel style={font=\scriptsize},
    xlabel={Leaderboard rank}, xlabel style={font=\scriptsize},
    xtick={1,5,9,13,17,21,25}, ytick={0,25,50,75,100},
    tick label style={font=\scriptsize},
    axis lines*=left,
    ymajorgrids, major grid style={black!10},
]
\addplot[fill=black!13, draw=none] coordinates {(1,90.69) (2,87.91) (3,86.95) (4,84.50) (5,74.62) (6,72.50) (7,64.83) (8,62.76) (10,61.18) (11,60.27) (12,60.03) (13,58.54) (14,58.35) (15,56.33) (16,54.80) (17,54.37) (18,49.14) (19,48.08) (20,42.03) (21,37.86) (22,30.61) (23,22.26) (24,21.74) (25,16.89) (26,5.66) (27,0.19)};
\addplot[fill=evON!60, draw=none] coordinates {(9,62.28)};
\node[font=\sffamily\tiny\bfseries, anchor=south, text=inkdark] at (axis cs:9,63) {ours, 62.28};
\end{axis}
\end{tikzpicture}
\caption{\textbf{Final public leaderboard}~\cite{eventbenchchallenge}. Average score of all 27 submissions; rank 9
(highlighted, 62.28\%) is this submission. A top cluster of four entries (84.50--90.69\%) is
separated from the rest by the largest gap ($84.50\%\to74.62\%$); the last three entries
(16.89\% to 0.19\%) may be incomplete submissions.}
\label{fig:leaderboard}
\end{figure}

Our rank (Figure~\ref{fig:leaderboard}) should be interpreted in light of our
training-free setting. Unlike approaches that fine-tune a model such as
EventGPT~\cite{eventgpt} on challenge training data, we use none of those data
and do not update the model weights. Consequently, leaderboard differences
need not reflect the quality of event representations alone.

\subsection{Analysis: What the Model Sees}
\label{sec:analysis}
Because test labels are withheld, we analyze our system's inputs and outputs instead of per-sample errors, asking two questions. \textbf{Q1:} what did the graded pipeline actually show the model? \textbf{Q2:} are its answers distributed as expected?

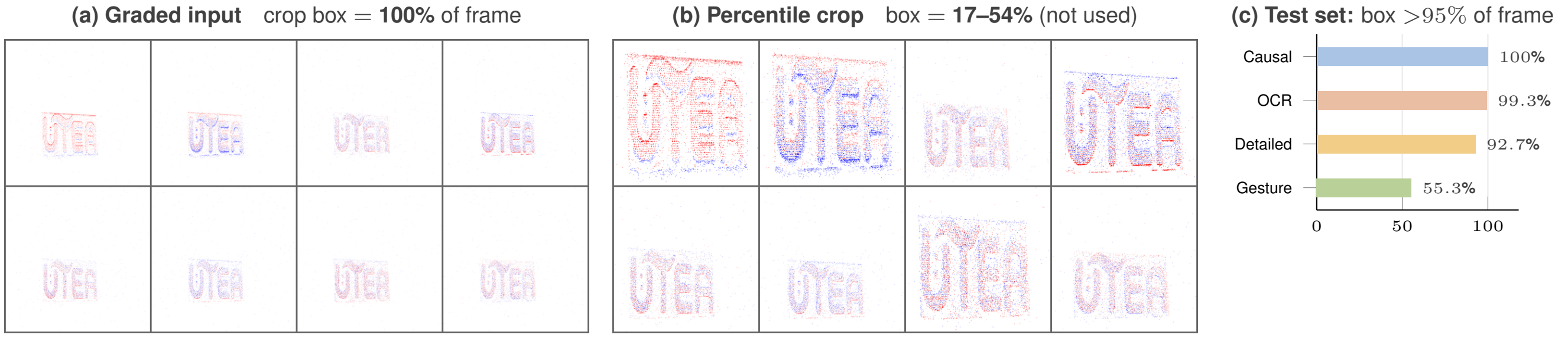
\begin{figure*}[t]
\centering
\begin{minipage}[t]{0.375\textwidth}\centering\vspace{0pt}
{\sffamily\scriptsize\color{inkdark}\textbf{(a) Graded input}\quad crop box $=$ \textbf{100\%} of frame}\\[3pt]
\includegraphics[width=\linewidth]{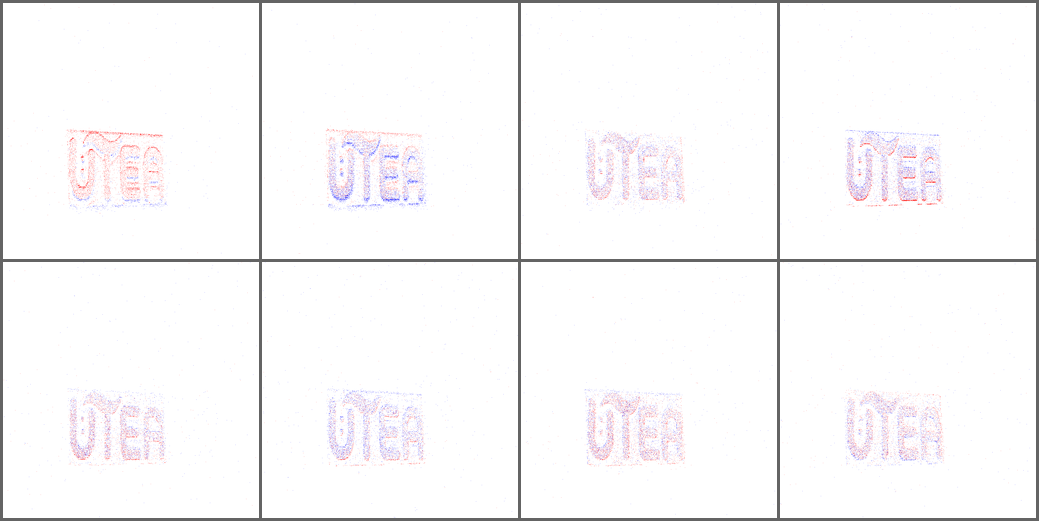}
\end{minipage}\hfill
\begin{minipage}[t]{0.375\textwidth}\centering\vspace{0pt}
{\sffamily\scriptsize\color{inkdark}\textbf{(b) Percentile crop}\quad box $=$ \textbf{17--54\%} (not used)}\\[3pt]
\includegraphics[width=\linewidth]{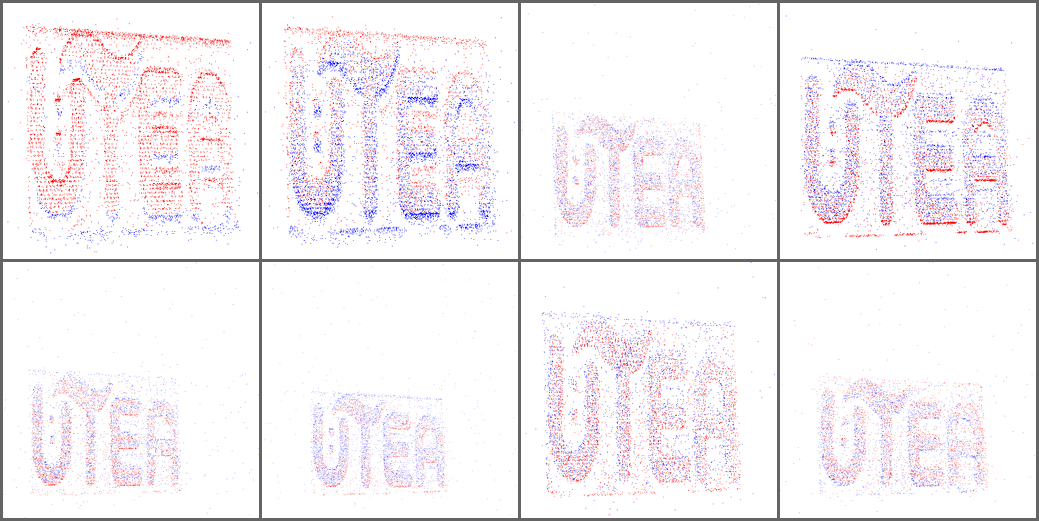}
\end{minipage}\hfill
\begin{minipage}[t]{0.22\textwidth}\centering\vspace{0pt}
{\sffamily\scriptsize\color{inkdark}\textbf{(c) Test set:} box ${>}95\%$ of frame}\\[1pt]
\begin{tikzpicture}
\begin{axis}[
    xbar, bar width=6pt, width=\linewidth, height=3.55cm,
    xmin=0, xmax=118, enlarge y limits=0.17,
    symbolic y coords={Gesture,Detailed,OCR,Causal}, ytick={Gesture,Detailed,OCR,Causal},
    y tick label style={font=\sffamily\tiny}, xtick={0,50,100}, x tick label style={font=\sffamily\tiny},
    axis lines*=left, xmajorgrids, major grid style={black!8}, bar shift=0pt,
    nodes near coords={\pgfmathprintnumber[fixed,precision=1]{\pgfplotspointmeta}\%},
    nodes near coords style={font=\sffamily\tiny\bfseries, text=inkdark, anchor=west},
    point meta=x,
]
\addplot[fill=catCR!80, draw=none] coordinates {(100,Causal)};
\addplot[fill=catOCR!80, draw=none] coordinates {(99.3,OCR)};
\addplot[fill=catDU!80, draw=none] coordinates {(92.7,Detailed)};
\addplot[fill=catGR!80, draw=none] coordinates {(55.3,Gesture)};
\end{axis}
\end{tikzpicture}
\end{minipage}
\caption{\textbf{Min/max cropping is defeated by noise.} (a) What our graded pipeline sends the
model for a public event OCR recording (ground truth ``UTEA''): a few isolated noise events reach
the border of every bin, so each ``tight'' crop spans the whole $1280{\times}720$ frame and the
text is heavily downsampled. (b) The same 8 bins with a 1st--99th percentile box isolate the text
on this recording, though not on most test recordings. Both use $256{\times}256$ cells, as in the
submission. (c) Share of all graded crop boxes on the test set covering more than 95\% of the
canvas; for gesture, the remaining boxes only trim unused canvas on lower-resolution clips.
\textbf{The crop almost never removed background.}}
\label{fig:ocrcrop}
\end{figure*}

\paragraph{Q1: the crop almost never tightened.} After the challenge we measured
every crop box the graded pipeline produced on the full test set. For event OCR, 99.3\% of the
1{,}260 per-bin boxes cover more than 95\% of the $1280{\times}720$ canvas, because a few
isolated noise events reach the sensor border in almost every bin. The same holds for
92.7\% of the 3{,}864 detailed-understanding boxes and for all 104 causal-reasoning boxes.
For gesture recognition, the shared box is smaller than half the canvas for exactly the 400
clips recorded below the $640{\times}480$ canvas size, and covers more than 95\% of it for
the other 494: the crop removed unused canvas, never background around the hand. The
practical consequence is largest for event OCR (Figure~\ref{fig:ocrcrop}(a)): each full
$1280{\times}720$ bin is squeezed into a $256{\times}256$ cell, a $5\times$ horizontal and
$2.8\times$ vertical downsampling that also distorts the aspect ratio of every character.

A robust box from the 1st--99th percentile of event coordinates isolates the text on some
recordings (Figure~\ref{fig:ocrcrop}(b)), but it is not a general fix: on the test set it
still covers more than 95\% of the canvas for 55\% of OCR bins (median 98\%), because in
many recordings the text or a textured background genuinely fills the frame. For event OCR
the binding constraint is therefore resolution rather than cropping, which points to
aspect-preserving cells, fewer and larger cells, or tiling as next steps. None of these
variants was used or scored in the graded submission.

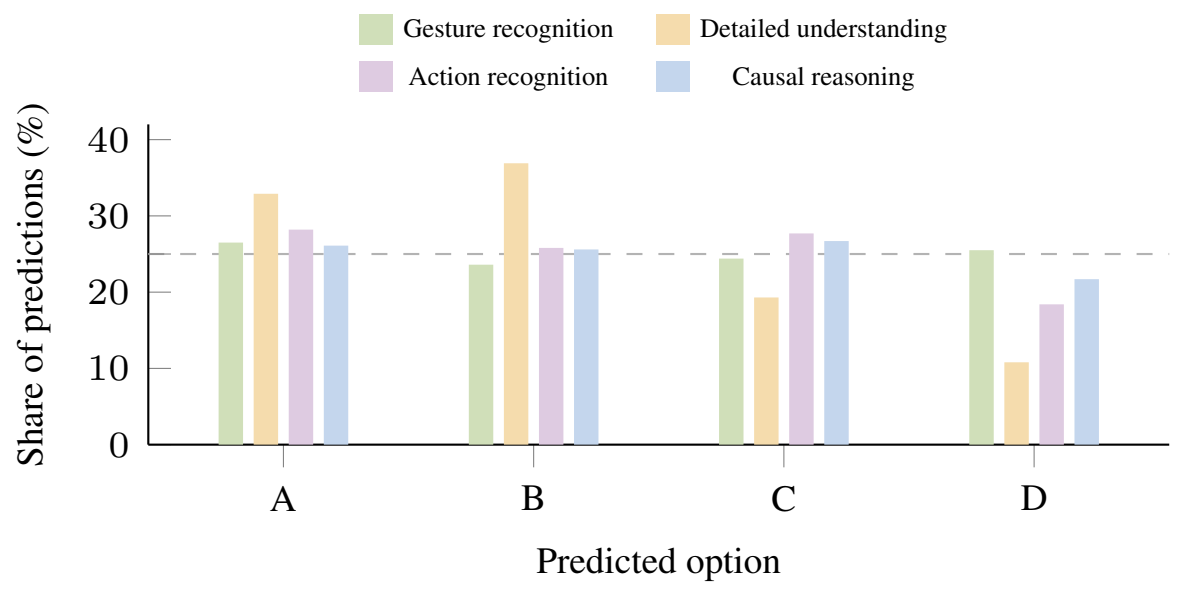
\begin{figure}[t]
\centering
\begin{tikzpicture}
\begin{axis}[
    width=\linewidth, height=3.7cm,
    ybar, bar width=4.6pt,
    ymin=0, ymax=42,
    ylabel={Share of predictions (\%)}, ylabel style={font=\scriptsize},
    symbolic x coords={A,B,C,D}, xtick=data,
    xlabel={Predicted option}, xlabel style={font=\scriptsize},
    tick label style={font=\scriptsize},
    enlarge x limits=0.18,
    axis lines*=left,
    extra y ticks={25}, extra y tick labels={},
    extra y tick style={grid=major, major grid style={dashed, black!30}},
    legend style={font=\tiny, draw=none, fill=none, at={(0.5,1.03)}, anchor=south, legend columns=2,
                  /tikz/every even column/.append style={column sep=6pt}},
    legend image code/.code={\draw[#1, draw=none] (0cm,-0.08cm) rectangle (0.22cm,0.12cm);},
]
\addplot[fill=catGR!55, draw=none] coordinates {(A,26.5) (B,23.6) (C,24.4) (D,25.5)};
\addplot[fill=catDU!55, draw=none] coordinates {(A,32.9) (B,36.9) (C,19.3) (D,10.8)};
\addplot[fill=catAR!55, draw=none] coordinates {(A,28.2) (B,25.8) (C,27.7) (D,18.4)};
\addplot[fill=catCR!55, draw=none] coordinates {(A,26.1) (B,25.6) (C,26.7) (D,21.7)};
\legend{Gesture recognition, Detailed understanding, Action recognition, Causal reasoning}
\end{axis}
\end{tikzpicture}
\caption{\textbf{Predicted answer letters.} Share of each letter in our predictions per multiple-choice category; the
dashed line marks a uniform 25\%. Detailed-understanding predictions are skewed toward A and
B.}
\label{fig:letters}
\end{figure}

\paragraph{Q2: answer-letter bias.} Without ground truth we cannot compute per-sample errors, but
the distribution of predicted letters is still informative (Figure~\ref{fig:letters}). For
gesture recognition, action recognition, and causal reasoning it stays close to uniform.
For detailed understanding, however, 69.8\% of predictions are A or B and only 10.8\% are D.
EventBench reports a near-uniform distribution of correct letters across the benchmark
(28.6\%, 24.7\%, 21.0\%, and 25.7\% for A--D~\cite{eventbench}); if the detailed-understanding
split follows it, the model under-predicts D by roughly half, a positional preference that
option shuffling or answer calibration could reduce without any training. Only one of the
1{,}922 multiple-choice predictions is not a valid letter (a detailed-understanding answer
that repeats an option's full text).

\subsection{Alternative Representations}
\label{sec:additional-experiments}
During development we scored several alternative representations and bin-selection
strategies directly against the evaluation server, each submitted for a single category.
All were run with Qwen3.5 9B. None outperformed the configuration finally used for its
category, so they were not adopted; we report them for completeness
(Table~\ref{tab:additional-experiments}; each technique is described in Appendix~\ref{app:impl}).

\begin{table}[t]
\centering
\footnotesize
\setlength{\tabcolsep}{3.4pt}
\newcommand{\dn}[1]{\textcolor{evON!85!black}{$-$#1}}
\begin{tabular}{@{}llrrr@{}}
\toprule
\textbf{Alternative (Qwen3.5 9B)} & \textbf{Category} & \textbf{Score} & \textbf{Adopted} & $\boldsymbol{\Delta}$ \\
\midrule
\rowcolor{catOCR!22} Foveated accumulation~\cite{foveated} & OCR & 29.63 & 34.57 & \dn{4.94} \\
\rowcolor{catOCR!22} Accumulation, activity & OCR & 21.60 & 34.57 & \dn{12.97} \\
\rowcolor{catOCR!22} Accumulation, signed & OCR & 22.22 & 34.57 & \dn{12.35} \\
\rowcolor{catCR!22} Category-aware selection & Causal & 55.00 & 77.22 & \dn{22.22} \\
\rowcolor{catDU!22} All above threshold & Detailed & 53.83 & 66.46 & \dn{12.63} \\
\rowcolor{catGR!22} MMR bin selection & Gesture & 49.47 & 54.70 & \dn{5.23} \\
\bottomrule
\end{tabular}
\caption{\textbf{Alternatives scored during development} (single-category submissions to the
evaluation server, accuracy in \%), against the configuration we adopted
(Table~\ref{tab:results}). The alternatives and the adopted grid-based configurations
use the same Qwen3.5 9B backbone. \textbf{No alternative beat the adopted configuration.}}
\label{tab:additional-experiments}
\end{table}

The three event OCR alternatives all collapse multiple bins into a single frame and score
21.60--29.63\%, consistent with our qualitative observation that strokes from different
moments become ambiguous once merged. Their comparison with the adopted 34.57\% grid
configuration is made using the same 9B backbone, eliminating model size as an explanation
for the gap. However, the configurations also differ in aggregation and preprocessing, so
these results do not isolate the effect of grid stitching itself. MMR bin selection was also
tried for detailed understanding, but we did not retain a comparable score for that run.

\section{Limitations and Future Work}
\label{sec:limits}

\paragraph{Ineffective cropping and low OCR resolution.} As shown in Section~\ref{sec:analysis},
min/max crop boxes were inflated to nearly the full canvas by isolated noise events on
almost every test clip, so the shared vs.\ per-bin distinction was largely
nominal and event OCR text was heavily downsampled. Robust cropping helps only on some
recordings; aspect-preserving or larger cells are the more promising, still training-free,
direction, and none has been scored yet.

\paragraph{Non-isolated OCR representation comparison.} Although the adopted OCR grid
and its scored alternatives all use Qwen3.5 9B, the alternatives change event aggregation
and preprocessing together. We therefore cannot attribute the score gap to grid stitching
alone without controlled, single-factor ablations.

\paragraph{No ground truth for analysis.} Because test labels are withheld, our post-hoc
analyses (crop boxes, answer-letter distribution, qualitative examples) describe the inputs
and outputs of our system but cannot attribute individual errors.

\paragraph{Unreconciled prediction merges.} Long batch runs needed many retry rounds after
transient endpoint failures and session limits, merged per category by a non-empty-wins
rule. For gesture recognition and detailed understanding, more complete merges were
produced after the graded submission but never resubmitted before the test server closed.

\paragraph{A stable failure subset.} Roughly 30--46 gesture recognition and detailed
understanding samples failed to produce a valid answer across every retry (5+ retries each,
across independent runs), suggesting degenerate event data for those samples rather than
transient failures; we did not visually confirm this.

\paragraph{Action recognition was not revisited.} EventGPT-Plus-2B was used in an earlier
project phase for an initial full-category submission. Its action recognition predictions
were never replaced by the grid pipeline and remain what was graded; this was not a
deliberate comparison, so we cannot say how the grid pipeline would have fared there.

\section{Conclusion}
\label{sec:conclusion}
A training-free pipeline, consisting of per-bin event rendering, task-specific bin
selection, and a single stitched grid queried with category-specific prompts, reaches 62.28\% overall
accuracy across five event-camera understanding tasks without using any of the challenge's
training data, with causal reasoning (77.22\%) showing that a frozen RGB-pretrained model
can reason over event renderings it was never trained on. Inspecting the exact model inputs
across the full test set points to concrete next steps: finer temporal sampling for
gestures, higher effective resolution for event OCR, where text was squeezed to a fraction
of each cell, and answer-order calibration for detailed understanding. All scored models
remain within the challenge's 10B parameter cap, with the same frozen 9B model used
across the four grid-based tasks.

\section*{Acknowledgments}
We thank the EventBench and EBMV @ ECCV 2026 organizers for releasing the benchmark, the
EventGPT-Plus-2B checkpoint, and the evaluation server.

\renewcommand{\bibfont}{\small}
\bibliographystyle{ieeenat_fullname}
\bibliography{refs}

\clearpage
\maketitlesupplementary
\appendix
\section{Exact Prompts}
\label{app:prompts}
Since prompt wording is the primary lever in this work rather than an implementation
detail, we give the exact system prompt for each grid-pipeline category, verbatim. For
gesture recognition, detailed understanding, causal reasoning, and event OCR, this is the
prompt that produced that category's graded predictions; the three multiple-choice prompts
share the same \code{Answer: X} instruction so a single extractor works across categories.
We also include the grid-pipeline prompt we tried for action recognition before settling on
EventGPT-Plus-2B (Section~\ref{sec:action}); it did not produce the graded predictions. Note
that the detailed understanding and event OCR prompts describe each cell as tightly
cropped, which Section~\ref{sec:analysis} shows was usually not the case.

\begin{tcolorbox}[breakable, colback=gray!3, colframe=gray!25, boxrule=0.4pt,
arc=4pt, left=4pt, right=4pt, top=3pt, bottom=3pt]
\scriptsize\ttfamily
\colorbox{catGR!40}{\parbox{\dimexpr\linewidth-2\fboxsep\relax}{\normalfont\scriptsize\bfseries\color{inkdark}\raggedright gesture\_recognition:}}\\
Event-camera frame grid, time order (left$\to$right, top$\to$bottom). Red=ON, Blue=OFF, White=no change. Frames cropped to the hand.\\
Judge the gesture from the hand's shape/position across frames: wave=side-to-side, clap=hands converge/diverge, rotation=circular, open/close=fingers spread/close, static=fixed pose.\\
End with exactly: Answer: X

\vspace{5pt}\normalfont\colorbox{catDU!40}{\parbox{\dimexpr\linewidth-2\fboxsep\relax}{\normalfont\scriptsize\bfseries\color{inkdark}\raggedright detailed\_understanding:}}\\
\ttfamily Event-camera frame grid, time order (left$\to$right, top$\to$bottom). Red=ON, Blue=OFF, White=no change. Each frame is cropped tightly to its own active region and may show a different zoom level than other frames --- don't assume matching scale.\\
Briefly note the objects/motion you see across frames, then answer using that.\\
End with exactly: Answer: X

\vspace{5pt}\normalfont\colorbox{catCR!40}{\parbox{\dimexpr\linewidth-2\fboxsep\relax}{\normalfont\scriptsize\bfseries\color{inkdark}\raggedright causal\_reasoning:}}\\
\ttfamily You are analyzing event camera frames to reason about cause and effect.\\
The grid shows frames in STRICT TIME ORDER: left$\to$right, then next row left$\to$right.\\
Red = ON event, Blue = OFF event, White = no change. Frames are cropped to the active region.\\
Consider what happens EARLY vs LATE in the sequence to identify causal relationships.\\
End your response with exactly: Answer: X (where X is A, B, C, or D)

\vspace{5pt}\normalfont\colorbox{catAR!40}{\parbox{\dimexpr\linewidth-2\fboxsep\relax}{\normalfont\scriptsize\bfseries\color{inkdark}\raggedright action\_recognition (experimental grid pipeline, not used for graded predictions; Section~\ref{sec:action}):}}\\
\ttfamily You are analyzing event camera frames showing human action.\\
The grid shows frames in STRICT TIME ORDER: left$\to$right, then next row left$\to$right.\\
Red = ON event, Blue = OFF event, White = no change.\\
Identify the action by how the body moves across the temporal sequence.\\
End your response with exactly: Answer: X (where X is A, B, C, or D)

\vspace{5pt}\normalfont\colorbox{catOCR!40}{\parbox{\dimexpr\linewidth-2\fboxsep\relax}{\normalfont\scriptsize\bfseries\color{inkdark}\raggedright event\_ocr (Qwen3.5 9B, Section~\ref{sec:inference}):}}\\
\ttfamily You are an expert OCR system reading text from event camera frames.\\
The grid shows separate snapshots of the same text region at different moments --- each cell only captures the edges that moved during that instant, so no single cell may show complete letters. Each cell is cropped tightly to its own active region and may show a different zoom level than other cells --- don't assume matching scale, just cross-reference the strokes themselves. Cross-reference strokes across ALL cells to reconstruct each character; a stroke missing in one cell may be clearly visible in another.\\
Red/Blue strokes trace the shape of letters and characters. White = no activity (background).\\
Read the text formed by combining the stroke patterns across cells.\\
Output ONLY the recognized text. Preserve capitalization, spacing, punctuation.\\
Multiple lines of text $\to$ output each on a separate line.\\
Do NOT describe the image or explain your reasoning.
\end{tcolorbox}

\section{Alternative Techniques}
\label{app:impl}
The alternatives scored in Table~\ref{tab:additional-experiments} work as follows.
\begin{itemize}[leftmargin=*, itemsep=1pt, topsep=2pt]
\item \textbf{Foveated accumulation}~\cite{foveated}: selects the busiest bins by event
count, sums their ON/OFF events into a \emph{single} frame rather than a grid, crops that
frame to the bounding box of the accumulated activity, log-normalizes, and upscales it.
\item \textbf{Accumulation, activity mode}: sums all events per pixel across the selected
bins irrespective of polarity, producing one grayscale event-density frame.
\item \textbf{Accumulation, signed mode}: sums polarity-signed events per pixel ($+1$ ON,
$-1$ OFF) into one frame around a gray midpoint (bright = ON-dominant, dark =
OFF-dominant).
\item \textbf{Category-aware bin selection}: routes each category to one of five candidate
strategies (event count, a coverage/redundancy/polarity quality score, spatial entropy,
temporal uniform, or MMR) via a lookup table. For causal reasoning the router
selected temporal uniform selection, the strategy already in use, so this run isolates the
effect of the routing implementation rather than a different selector.
\item \textbf{All above threshold}: as described in Section~\ref{sec:binsel}, keeps every
bin above a data-driven event-count floor, so the grid grows or shrinks with the number of
bins that pass.
\item \textbf{MMR bin selection}: a greedy maximal-marginal-relevance (MMR)~\cite{mmr} selector that picks
the bin maximizing $\lambda \cdot \text{quality} - (1-\lambda) \cdot
\text{similarity to bins already picked}$ ($\lambda{=}0.5$), with quality from coverage,
redundancy, and polarity balance, and similarity computed on downsampled event-density
maps.
\end{itemize}

\section{Additional Test-Set Statistics}
\label{app:stats}
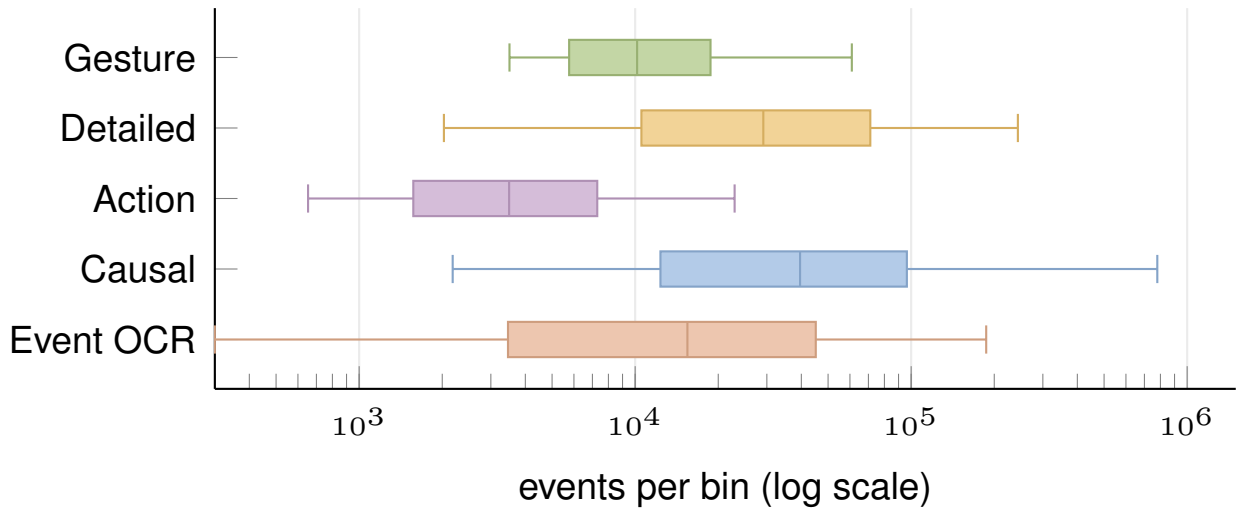
\begin{figure}[h]
\centering
\begin{tikzpicture}
\begin{axis}[
    width=\linewidth, height=4.1cm, xmode=log, xmin=300, xmax=1.5e6,
    ytick={1,2,3,4,5}, yticklabels={Event OCR,Causal,Action,Detailed,Gesture},
    y tick label style={font=\sffamily\scriptsize}, x tick label style={font=\sffamily\tiny},
    xlabel={events per bin (log scale)}, xlabel style={font=\sffamily\scriptsize},
    axis lines*=left, xmajorgrids, major grid style={black!8},
    boxplot/draw direction=x, boxplot/box extend=0.5,
]
\addplot+[fill=catGR!70, draw=catGR!90!black, mark=none, boxplot prepared={lower whisker=3505, lower quartile=5765, median=10164, upper quartile=18752, upper whisker=60965, draw position=5}] coordinates {};
\addplot+[fill=catDU!70, draw=catDU!90!black, mark=none, boxplot prepared={lower whisker=2028, lower quartile=10535, median=29132, upper quartile=71126, upper whisker=243580, draw position=4}] coordinates {};
\addplot+[fill=catAR!70, draw=catAR!90!black, mark=none, boxplot prepared={lower whisker=653, lower quartile=1571, median=3494, upper quartile=7290, upper whisker=22928, draw position=3}] coordinates {};
\addplot+[fill=catCR!70, draw=catCR!90!black, mark=none, boxplot prepared={lower whisker=2182, lower quartile=12349, median=39588, upper quartile=96484, upper whisker=779646, draw position=2}] coordinates {};
\addplot+[fill=catOCR!70, draw=catOCR!90!black, mark=none, boxplot prepared={lower whisker=300, lower quartile=3462, median=15465, upper quartile=45123, upper whisker=187035, draw position=1}] coordinates {};
\end{axis}
\end{tikzpicture}
\caption{\textbf{Events per bin by category}: boxes span the 25th to 75th percentile, whiskers
the 5th to 95th; the event OCR whisker is clipped at the axis because its 5th percentile is an
empty bin.}
\label{fig:evbins}
\end{figure}

Table~\ref{tab:suppstats} extends Table~\ref{tab:data} with the file-name groups inside each
category, a proxy for the underlying source datasets since the release has no per-file source
labels, and with the spread of events per bin, which Figure~\ref{fig:evbins} plots on a log
scale. Density differs by more than an order of magnitude across categories. The median
causal-reasoning bin holds about 11 times as many events as the median action-recognition bin
(39.6k vs.\ 3.5k), and the densest 5\% of causal-reasoning bins exceed 780k events, about ten
per pixel of a $320{\times}240$ frame, which renders as an almost uniformly colored cell. At
the other end, 15.6\% of event OCR bins are empty. A single fixed rendering therefore receives
inputs ranging from almost blank to saturated, which motivates density-adaptive rendering as
future work.

\begin{table}[h]
\centering
\scriptsize
\setlength{\tabcolsep}{2.6pt}
\begin{tabular}{@{}l >{\raggedright\arraybackslash}p{2.35cm} rrrrr@{}}
\toprule
& & \multicolumn{5}{c}{\textbf{Events per bin (percentile)}} \\
\cmidrule(l){3-7}
\textbf{Category} & \textbf{File-name groups (clips)} & \textbf{5} & \textbf{25} & \textbf{50} & \textbf{75} & \textbf{95} \\
\midrule
\rowcolor{catGR!22} Gesture & camera* 494, X* 400 & 3.5k & 5.8k & 10.2k & 18.8k & 61.0k \\
\rowcolor{catDU!22} Detailed & video-id 300, v\_* 45, X* 7, other 131 & 2.0k & 10.5k & 29.1k & 71.1k & 243.6k \\
\rowcolor{catAR!22} Action & C\#P\#* 220, dvSave* 145 & 0.7k & 1.6k & 3.5k & 7.3k & 22.9k \\
\rowcolor{catCR!22} Causal & v\_* 104 & 2.2k & 12.3k & 39.6k & 96.5k & 779.6k \\
\rowcolor{catOCR!22} Event OCR & recording* 162 & 0 & 3.5k & 15.5k & 45.1k & 187.0k \\
\bottomrule
\end{tabular}
\caption{\textbf{Test-set composition and event density}, measured on the released files
(4{,}712 to 16{,}760 bins per category).}
\label{tab:suppstats}
\end{table}

\section{Event OCR Outputs}
\label{app:ocrout}
Of our 162 graded event OCR outputs, 124 contain Chinese characters, 37 contain only Latin
script, and one is a single digit. 46 outputs span more than one line, the median output is 7
characters long (the longest is 68), and none is empty. As disclosed in
Section~\ref{sec:inference}, these graded predictions come from the Qwen3.5 9B OCR
pipeline; repeated inference runs were merged by retaining non-empty answers as
described in Section~\ref{sec:impl}.

\section{More Test Samples}
\label{app:examples}
The cards below show two further test samples per category, drawn uniformly at random with a
fixed seed (2026) from the test set, excluding the samples of Figure~\ref{fig:examples};
nothing was filtered or re-drawn. They are rendered as in Figure~\ref{fig:examples}, and the
filled circle again marks our prediction, not the ground truth. The two event OCR samples
illustrate the range of behavior. In the first, the model reproduces the Chinese title and
transliterates the Russian line below it into Latin letters. In the second, the visible text
is Chinese, yet the model returns an unrelated English phrase (``Welcome to the Event!''), a
hallucination that free-text OCR is exposed to and multiple-choice tasks are not.

\par\medskip
{\renewcommand{\exsep}{\par\vspace{7pt}}%
\excard{catGR}{Gesture recognition}{\includegraphics[width=\linewidth]{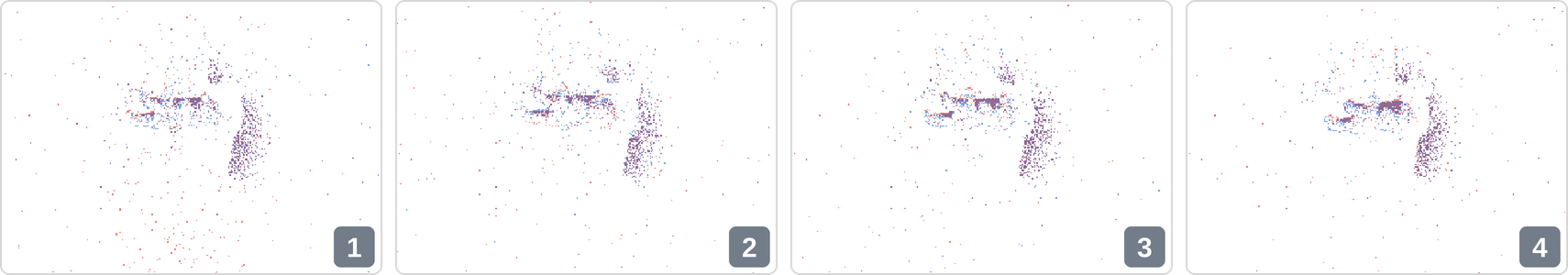}}{%
\exq{Are hands forming any specific gestures in this scene?}
\exopts{%
\exopt{A}{Finger Tapping}
\exopt{B}{Nose-reaching maneuver}
\exopt{C}{Tremor (hands outstretched)}
\exoptpred{D}{Hand movements (opening-closing)}
}
}{320$\times$240}{32}{Qwen3.5 9B}
\exsep
\excard{catGR}{Gesture recognition}{\includegraphics[width=\linewidth]{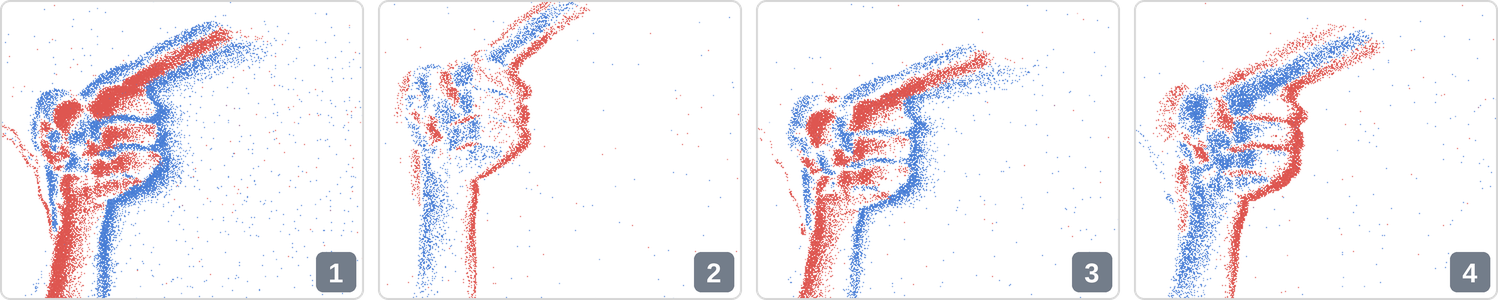}}{%
\exq{Are there any recognizable hand gestures in this scene?}
\exoptsgrid{\exopt{A}{hand waves}}{\exoptpred{B}{pointing}}{\exopt{C}{the 'scissors' sign}}{\exopt{D}{the 'rock' sign}}
}{640$\times$480}{8}{Qwen3.5 9B}
\exsep
\excard{catDU}{Detailed understanding}{\includegraphics[width=\linewidth]{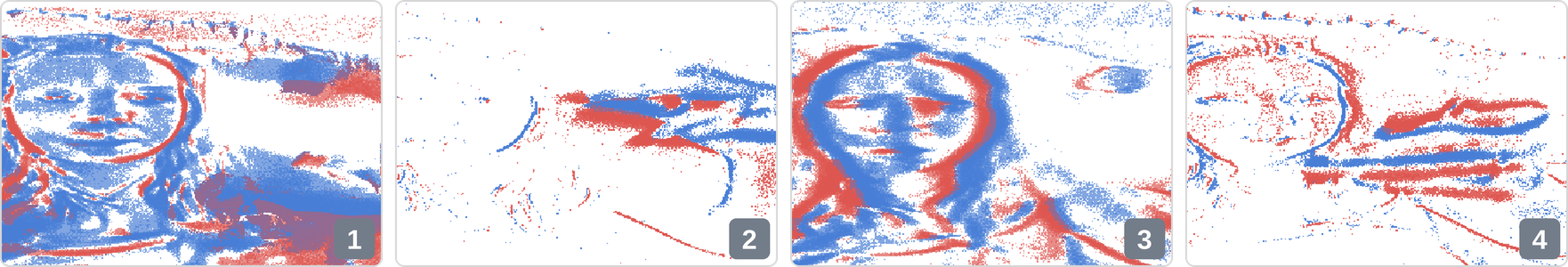}}{%
\exq{What does the woman do with the small green object throughout the video?}
\exopts{%
\exoptpred{A}{She holds it between her fingers and moves it slightly}
\exopt{B}{She spins it rapidly}
\exopt{C}{She drops it repeatedly}
\exopt{D}{She places it on a surface}
}
}{320$\times$240}{16}{Qwen3.5 9B}
\exsep
\excard{catDU}{Detailed understanding}{\includegraphics[width=\linewidth]{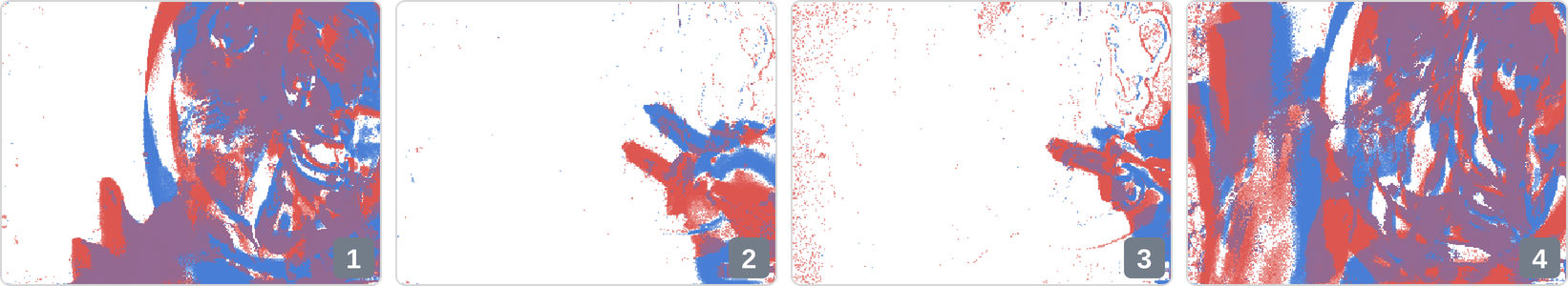}}{%
\exq{How does the wood-burning tool move across the wooden surface?}
\exopts{%
\exopt{A}{In circular motions}
\exopt{B}{Back and forth in straight lines}
\exopt{C}{Rapidly and erratically}
\exoptpred{D}{Slowly and precisely}
}
}{320$\times$240}{16}{Qwen3.5 9B}
\exsep
\excard{catAR}{Action recognition}{\includegraphics[width=\linewidth]{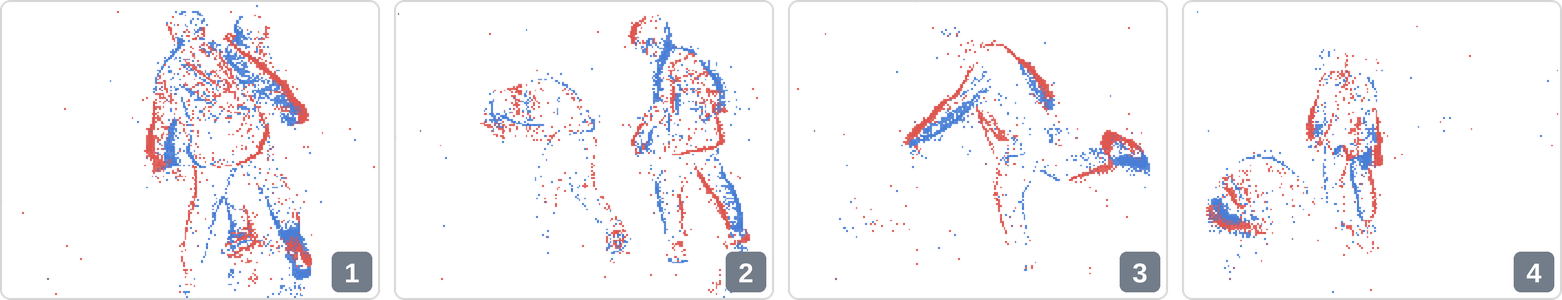}}{%
\exq{what is the most likely action?}
\exoptsgrid{\exopt{A}{walking}}{\exopt{B}{hairgrabs}}{\exopt{C}{strangling}}{\exoptpred{D}{kicking}}
}{346$\times$260}{16}{EventGPT-Plus-2B}
\exsep
\excard{catAR}{Action recognition}{\includegraphics[width=\linewidth]{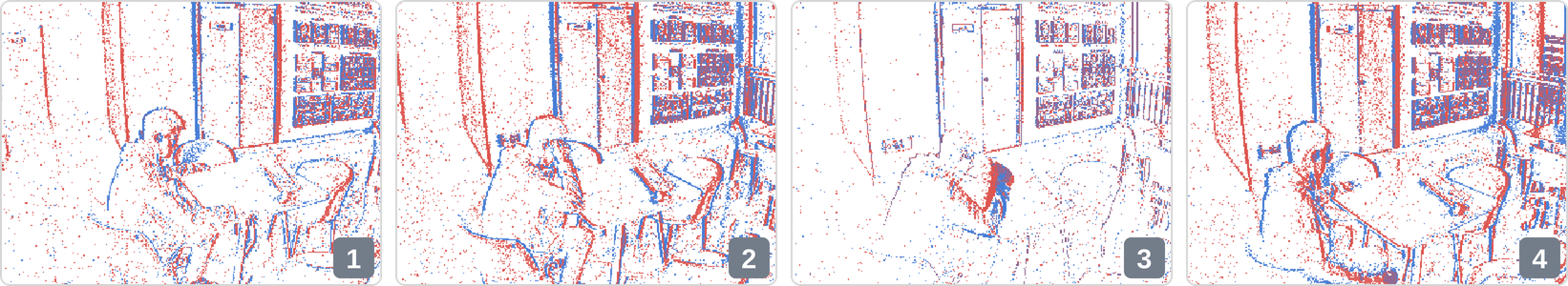}}{%
\exq{What is the main action being performed?}
\exoptsgrid{\exoptpred{A}{None of the above}}{\exopt{B}{exchange things}}{\exopt{C}{lie down}}{\exopt{D}{folding clothes}}
}{320$\times$240}{8}{EventGPT-Plus-2B}
\exsep
\excard{catCR}{Causal reasoning}{\includegraphics[width=\linewidth]{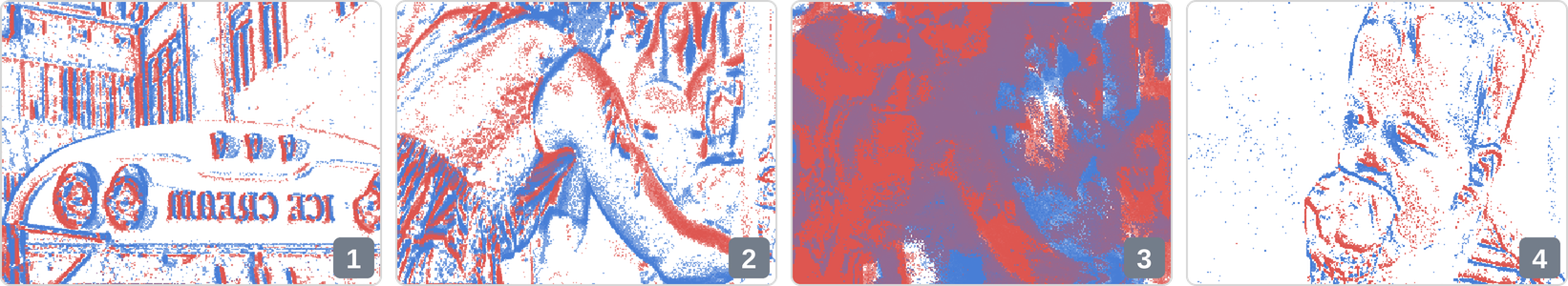}}{%
\exq{Why is the text 'ICE CREAM' on the front of the truck written in reverse, based on the truck's position and orientation in the scene?}
\exopts{%
\exopt{A}{To match the reversed reflection in the truck's own windshield.}
\exoptpred{B}{To be legible in a rear-view mirror of a vehicle ahead of the truck.}
\exopt{C}{So pedestrians approaching from the front can read it directly.}
\exopt{D}{As a decorative design unrelated to readability.}
}
}{320$\times$240}{16}{Qwen3.5 9B}
\exsep
\excard{catCR}{Causal reasoning}{\includegraphics[width=\linewidth]{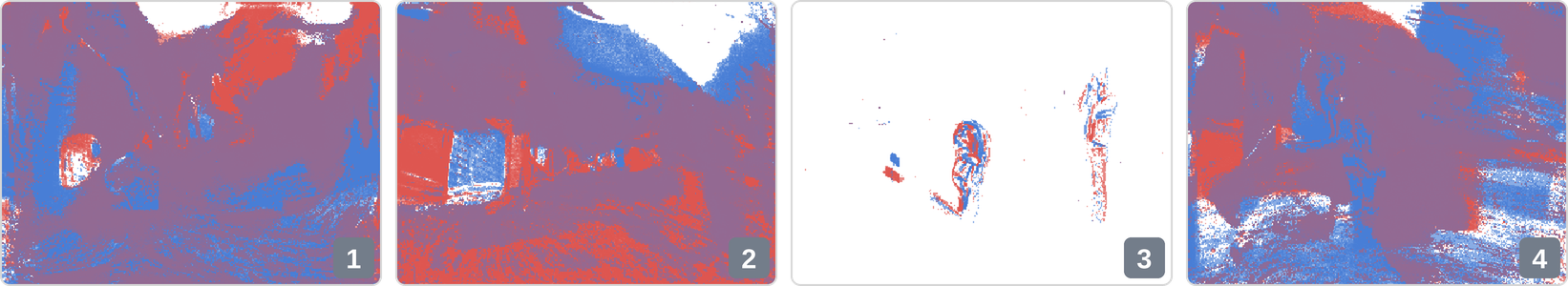}}{%
\exq{Why does the rake handle strike the person holding it in the head during the first incident?}
\exopts{%
\exopt{A}{The other person pushes the rake handle toward them.}
\exoptpred{B}{They step on the rake's head while it is lying flat on the ground.}
\exopt{C}{They swing the rake upward intentionally.}
\exopt{D}{The rake is pulled by a hidden string from off-screen.}
}
}{320$\times$240}{16}{Qwen3.5 9B}
\exsep
\excard{catOCR}{Event OCR}{\includegraphics[width=\linewidth]{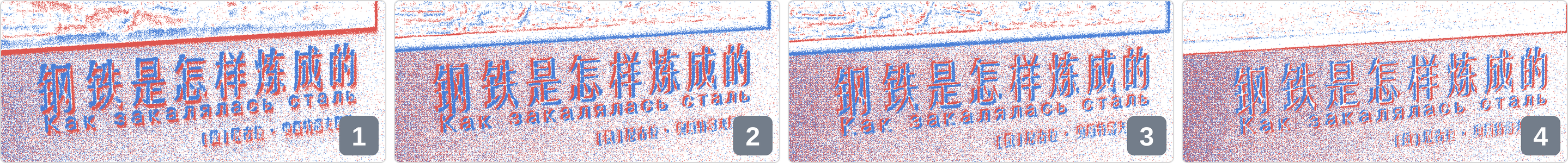}}{%
\exq{Recognize the text displayed and write it down. Only output the recognized text.}
\expredimg{catOCR}{\includegraphics[scale=0.2]{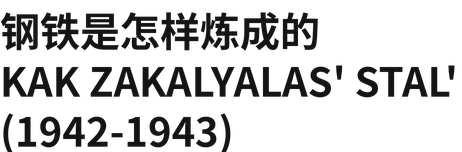}}
}{1280$\times$720}{16}{Qwen3.5 9B}
\exsep
\excard{catOCR}{Event OCR}{\includegraphics[width=\linewidth]{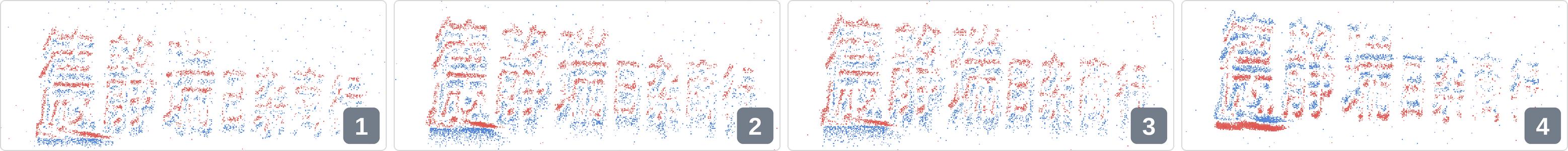}}{%
\exq{Scan the scene for text and return it. Only output the recognized text.}
\expredtxt{catOCR}{Welcome to the Event!}
}{1280$\times$720}{8}{Qwen3.5 9B}
}

\end{document}